\documentclass[11pt]{worldstream}

\usepackage{tabularx}
\usepackage{array}
\usepackage{float}
\usepackage{listings}
\usepackage{tikz}
\usetikzlibrary{arrows.meta,positioning,fit,calc}

\newcolumntype{Y}{>{\raggedright\arraybackslash}X}
\newcolumntype{C}{>{\centering\arraybackslash}X}
\newcommand{\system}{\textsc{WorldStream}}

\colorlet{youtuBlue}{reportpurple}
\colorlet{youtuLightBlue}{reportlightpurple}
\definecolor{youtuText}{HTML}{1C2B33}
\definecolor{gainRed}{HTML}{D5533D}
\definecolor{gainGreen}{HTML}{2F8A68}

\newcommand{\metricgain}[1]{\textcolor{gainGreen}{\sffamily\bfseries #1}}
\newcommand{\tablekey}[1]{\textcolor{reportdeeppurple}{\sffamily\bfseries #1}}
\newcommand{\tablescope}[1]{\textit{#1}}
\hypersetup{
  pdftitle={WorldStream: Streaming Interactive 3D Worlds with Hierarchical Planning and Persistent State},
  pdfsubject={Geography-aware planning, streaming execution, persistent state, and grounded interaction}}

\title{\textbf{WorldStream}: Streaming Interactive 3D Worlds with Hierarchical Planning and Persistent State}
\author{%
Xiaoxiao Ma\textsuperscript{1,2}\quad
Weiyang Jin\textsuperscript{2}\quad
Yanwen Ma\textsuperscript{2}\quad
Haoran Li\textsuperscript{2}\quad
Zhixiang Wei\textsuperscript{1}\\[-.1em]
Songchun Zhang\textsuperscript{2}\quad
Yaowei Li\textsuperscript{2}\quad
Junhao Zhuang\textsuperscript{2}\quad
Zeyue Xue\textsuperscript{2,\ensuremath{\dagger}}\quad
Yuming Li\textsuperscript{2}\\[-.1em]
Zheming Fu\textsuperscript{2}\quad
Haoyang Huang\textsuperscript{2}\quad
Nan Duan\textsuperscript{2}\quad
Feng Zhao\textsuperscript{1,\ensuremath{*}}}
\affiliation{%
\textsuperscript{1}University of Science and Technology of China\quad
\textsuperscript{2}Joy Future Academy, JD\\[-.1em]
\textsuperscript{\ensuremath{\dagger}}Project Lead\qquad
\textsuperscript{\ensuremath{*}}Corresponding Author: \email{fzhao956@ustc.edu.cn}}

\abstract{%
We present \textbf{\system}, an open-ended 3D world model with three simultaneous capabilities: it extends playable terrain beyond any pre-specified map, acts on open-ended language in the running world, and restores visited places with their edits. A hierarchical planner over a typed decision model emits structured decision state: Macro fixes settlement, ecology, transport, and hydrological structure across a group of tiles; Micro specializes that contract into tile semantics under measured terrain and accepted neighbor commitments. Deterministic local code compiles accepted plans into terrain, roads, settlements, vegetation, and entities. Streaming separates their reach: plans run ahead along the predicted corridor, compilation only where movement will soon need it, geometry behind the user released. One world record retains plans, seeds, object identities and mutable state, boundary commitments, and edits by attribute, revised as the user advances and edits; returning tiles recompile from it. The same record grounds interaction: the model acts on existing entities, including stateful NPCs that update as they act, edits the present region, or attaches intent to unplanned regions. Together they yield a world expanding under bounded budgets, connected across independently planned tiles, and interactive.%
}

\checkdata[Project and Demo]{\url{https://krennic999.github.io/WorldStream/}}
\checkdata[Blog]{\url{https://krennic999.github.io/WorldStream/blog/}}

\begin{document}
\maketitle

\begin{figure}[H]
\centering
\captionsetup{font=footnotesize}
\includegraphics[width=.96\textwidth]{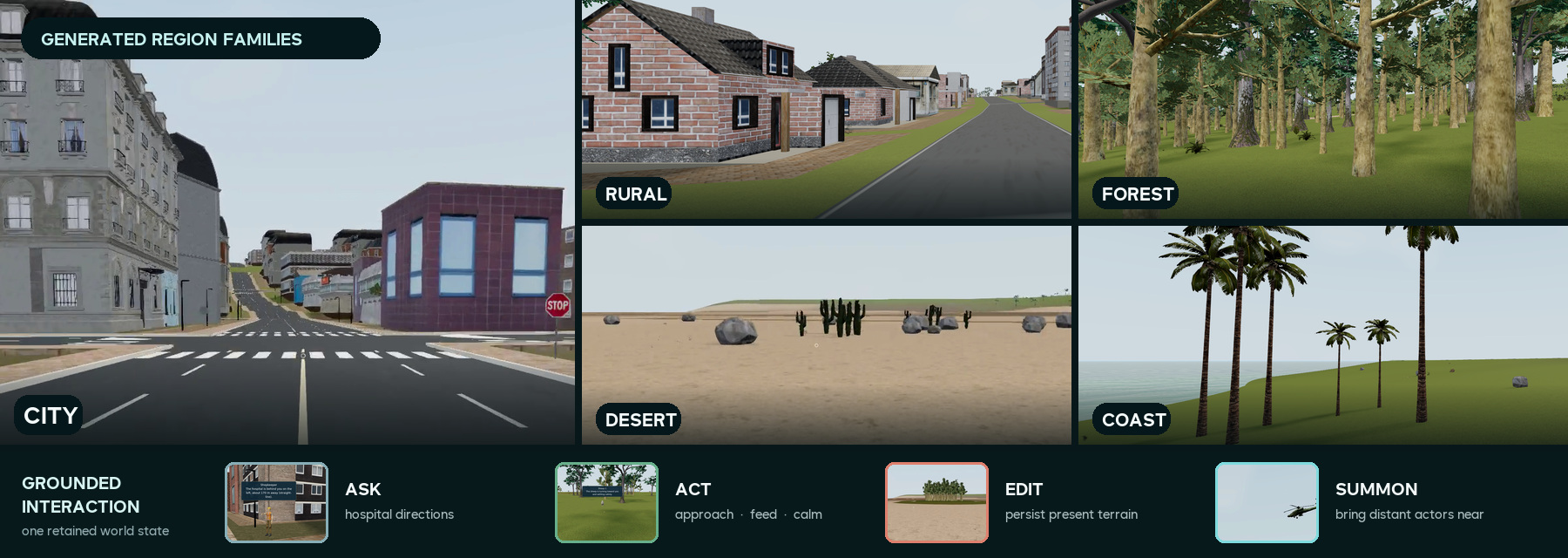}
\caption{Representative WorldStream v1 runtime regions include dense city blocks, a rural settlement, streamed forest, desert terrain, and coastal terrain (top). A shared retained-state interface grounds language actions in the running world: users can ask for directions, direct nearby actors, edit present terrain, and summon distant actors (bottom).}
\label{fig:teaser}
\end{figure}

\makeabstract

\section{Introduction}
\label{sec:introduction}

World models have moved from latent dynamics used for prediction and control~\cite{ha2018worldmodels,hafner2019planet,schrittwieser2020muzero,hafner2023dreamerv3} to interactive visual environments that a person can enter and act in~\cite{bruce2024genie,valevski2024gamengen,alonso2024diamond,zhen2024threedvla,qin2025worldsimbench,worldlabs2026atlas,liu2026r2wam,zhou2026wovrplus,chen2026h3world}. Open-world games, embodied simulation, and interactive content creation have all adopted this second form, and they impose a requirement that prediction alone does not address. A predictive model is evaluated on the view it produces next; an interactive environment is evaluated on what persists after that view is replaced. Its extent, contents, and accumulated user history must outlive any rendered frame.

Two lines of work each supply part of this. Learned generative world models produce responsive views from user action, and their consistency horizon has extended steadily~\cite{yang2023unisim,hu2023gaia,zhou2025persistent,sun2025worldplay}. Their state is maintained in frames and latents, providing no persistent place identifier: a region released from memory can only be resynthesized, and an individual object cannot be referenced for later modification. Procedural and code-driven 3D systems hold the complementary position~\cite{parish2001cities,deitke2022procthor,rais2023infinigen,xie2023citydreamer,li2024procgs,sun2023threedgpt,hu2024scenecraft,yang2023holodeck,tam2024scenemotifcoder,liu2026citygenagent,guo2026worldclaw,chen2026codeworldmodel}. Learned object generation and stylization complement these systems by producing reusable geometry from a reference image, object-part sequence, or text instruction~\cite{zhang2025ar,zhang2025tar3d,zhang2024temo}. Explicit scene systems compile geometry from semantic specifications over such asset libraries, leaving every object inspectable and editable, but pay their generation cost before play begins over a scene fixed in advance. That setting determines neither which part of an unbounded world to prepare next, nor what should persist once a prepared region is unloaded.

Open-ended exploration requires both at once, and imposes three concurrent requirements on the runtime. First, the world must extend past any pre-specified map while the user is inside it, under a per-frame cost that does not grow with explored extent. Second, open-ended language must be interpreted against the entities and terrain that currently exist and applied to the running world, covering both the region the user occupies and regions that have not yet been generated. Third, a previously visited place must be reproduced, together with its accepted edits, after its geometry has been evicted to keep render memory bounded. No single stage of a generation pipeline supplies all three: extent is a scheduling property, grounded change is a representation property, and reproducibility is a property of what the system stores.

We present \system{}, which meets these requirements by using a large decision model for planning and deterministic local code for realization, with both operating on a single persistent world record. Planning is organized at two complementary spatial scales. Macro planning takes regional terrain and water summaries together with the accepted state of nearby tiles as input, and determines multi-tile settlement, ecology, transport, and hydrological structure; Micro planning specializes each regional contract into executable tile semantics, under the measured surface and the boundary commitments its neighbors have already made. Both scales are required, as each addresses a distinct failure mode the other does not: a planner conditioned only on local evidence produces terrain-compatible but regionally unrelated tiles, whereas a regional plan alone does not determine how its structure continues across each local boundary. At both scales, Jev produces decisions under a typed schema in which every field returns a preferred value together with a distribution over the permitted alternatives~\cite{jev2026}. A bounded resolution procedure then searches those distributions for a joint assignment admitted by the terrain, the regional contract, and the already-accepted neighbors, which establishes that a plan is executable before any geometry is built.

To connect model reasoning with spatial construction, \system{} expresses every accepted plan as a compact, executable semantic specification. The specification records how a place is organized, which content families it contains, how it connects to neighboring tiles, and which parameters its generators receive. Deterministic local code expands this information into terrain, roads, settlements, vegetation, and entities, using canonical road ports and continuous height and water sampling to connect independently processed tiles. One model decision can therefore control many concrete objects while local code resolves their coordinates and geometric constraints deterministically. This explicit separation supports two streaming horizons: planning establishes semantics ahead of movement, and construction realizes the subset approaching the active area, keeping runtime work bounded as explored extent grows.

\begin{figure}[!t]
\centering
\resizebox{\textwidth}{!}{\begin{tikzpicture}[
  flow/.style={-{Latex[length=2.5mm,width=1.6mm]}, line width=.9pt, draw=workflowarrow},
  paneltitle/.style={font=\fontsize{8.4}{9.6}\selectfont\sffamily\bfseries,
    text=workflowink, align=center},
  note/.style={font=\fontsize{6.8}{7.9}\selectfont\sffamily\bfseries, align=center},
  tag/.style={draw=workflowborder, rounded corners=3pt, fill=white,
    inner xsep=3pt, inner ysep=1.5pt,
    font=\fontsize{5.5}{6.3}\selectfont\sffamily\bfseries, text=workflowink},
  edgelabel/.style={font=\fontsize{5.5}{6.3}\selectfont\sffamily,
    text=workflowarrow, fill=white, inner sep=1pt},
  iconline/.style={draw=workflowink, line width=.62pt, line join=round, line cap=round},
  fine/.style={draw=workflowink!70, line width=.42pt, line join=round, line cap=round}
]

\definecolor{workflowborder}{HTML}{D9DEE2}
\definecolor{workflowarrow}{HTML}{69737B}
\definecolor{workflowink}{HTML}{374047}
\definecolor{workflowlavender}{HTML}{EDE5F1}
\definecolor{workflowgreen}{HTML}{C8D8C2}
\definecolor{workflowsand}{HTML}{E3D5AE}
\definecolor{workflowblue}{HTML}{B9D8DF}
\definecolor{workflowslate}{HTML}{DDE3E7}
\definecolor{workflowcoral}{HTML}{C8735D}
\definecolor{workflowdarkgreen}{HTML}{4F7C68}

\node[paneltitle] at (2.00,3.35) {(a) World knowledge};
\node[paneltitle] at (6.00,3.35) {(b) Hierarchical planning};
\node[paneltitle] at (10.00,3.35) {(c) Local compilation};
\node[paneltitle] at (14.00,3.35) {(d) 3D realization};
\foreach \x in {2,6,10,14}{
  \draw[workflowborder,line width=.55pt] (\x-1.48,3.08) -- (\x+1.48,3.08);
}

\node[tag] at (1.24,2.75) {terrain};
\node[tag] at (2.00,2.75) {assets};
\node[tag] at (2.76,2.75) {rules};
\filldraw[iconline,fill=workflowslate]
  (1.02,1.45) -- (2.00,1.02) -- (2.98,1.45) -- (2.00,1.88) -- cycle;
\filldraw[iconline,fill=workflowsand]
  (1.02,1.74) -- (2.00,1.31) -- (2.98,1.74) -- (2.00,2.17) -- cycle;
\filldraw[iconline,fill=workflowgreen]
  (1.02,2.03) -- (2.00,1.60) -- (2.98,2.03) -- (2.00,2.46) -- cycle;
\draw[fine]
  (1.26,2.02) .. controls (1.52,2.24) and (1.75,1.83) .. (2.02,2.09)
  .. controls (2.28,2.33) and (2.47,1.93) .. (2.75,2.11);
\draw[workflowblue,line width=1.0pt]
  (1.73,1.75) .. controls (1.84,1.95) and (1.88,2.16) .. (2.10,2.40);
\fill[reportpurple] (1.42,2.04) circle (.055cm);
\fill[reportpurple] (2.47,1.99) circle (.055cm);
\draw[workflowink,line width=.55pt] (1.29,1.63) -- (1.53,1.52) -- (1.77,1.63);
\fill[workflowdarkgreen] (2.54,1.63) circle (.045cm);
\fill[workflowdarkgreen] (2.68,1.57) circle (.035cm);

\node[tag] at (5.27,2.75) {MACRO};
\node[tag] at (6.82,2.75) {MICRO};
\filldraw[iconline,fill=workflowlavender] (4.75,1.53) rectangle (5.78,2.56);
\foreach \x in {5.093,5.436}{\draw[fine] (\x,1.53) -- (\x,2.56);}
\foreach \y in {1.873,2.216}{\draw[fine] (4.75,\y) -- (5.78,\y);}
\fill[reportpurple!28] (5.436,1.873) rectangle (5.78,2.216);
\draw[reportpurple,line width=.8pt] (5.436,1.873) rectangle (5.78,2.216);
\draw[workflowink,line width=.85pt] (4.75,1.98) -- (5.78,1.98);
\draw[workflowink,line width=.85pt] (5.24,1.53) -- (5.24,2.56);
\fill[workflowdarkgreen] (4.91,2.38) circle (.035cm);
\fill[workflowdarkgreen] (5.63,1.70) circle (.035cm);
\filldraw[iconline,fill=white] (6.37,1.38) rectangle (7.27,2.46);
\draw[fine] (6.82,1.38) -- (6.82,2.46);
\draw[fine] (6.37,1.92) -- (7.27,1.92);
\fill[workflowgreen] (6.37,1.92) rectangle (6.82,2.46);
\fill[workflowsand] (6.82,1.38) rectangle (7.27,1.92);
\draw[reportpurple,dashed,line width=.55pt] (5.78,2.216) -- (6.37,2.46);
\draw[reportpurple,dashed,line width=.55pt] (5.78,1.873) -- (6.37,1.38);
\draw[workflowink,line width=.78pt] (6.37,1.92) -- (7.27,1.92);
\draw[workflowink,line width=.78pt] (6.82,1.38) -- (6.82,2.46);
\draw[workflowink,line width=.58pt] (6.48,2.02) -- (6.61,2.15) -- (6.72,2.26);
\filldraw[iconline,fill=workflowblue] (6.94,2.10) rectangle (7.13,2.32);
\fill[workflowdarkgreen] (6.53,2.16) circle (.045cm);
\fill[reportpurple] (6.95,1.66) circle (.06cm);

\filldraw[iconline,fill=workflowink] (8.58,1.39) rectangle (9.62,2.55);
\draw[white!45,line width=.45pt] (8.58,2.31) -- (9.62,2.31);
\fill[workflowsand] (8.73,2.43) circle (.035cm);
\fill[workflowgreen] (8.85,2.43) circle (.035cm);
\draw[workflowblue,line width=.7pt] (8.78,2.09) -- (9.38,2.09);
\draw[workflowgreen,line width=.7pt] (8.78,1.89) -- (9.48,1.89);
\draw[workflowsand,line width=.7pt] (8.78,1.69) -- (9.29,1.69);

\filldraw[iconline,fill=workflowblue] (9.87,2.31) circle (.115cm);
\draw[workflowink,line width=.5pt] (9.80,2.31) -- (9.94,2.31);
\filldraw[iconline,fill=workflowsand] (9.87,1.96) circle (.115cm);
\fill[workflowink] (9.82,1.91) rectangle (9.92,2.01);
\filldraw[iconline,fill=workflowgreen] (9.87,1.61) circle (.115cm);
\fill[workflowdarkgreen] (9.83,1.61) circle (.025cm);
\fill[workflowdarkgreen] (9.91,1.66) circle (.022cm);
\draw[fine] (9.62,2.08) -- (9.75,2.31);
\draw[fine] (9.62,1.96) -- (9.75,1.96);
\draw[fine] (9.62,1.83) -- (9.75,1.61);
\draw[fine] (9.99,2.31) -- (10.38,2.08);
\draw[fine] (9.99,1.96) -- (10.38,1.96);
\draw[fine] (9.99,1.61) -- (10.38,1.84);

\filldraw[iconline,fill=workflowgreen]
  (10.38,1.78) -- (10.82,1.56) -- (11.26,1.78) -- (10.82,2.00) -- cycle;
\filldraw[iconline,fill=workflowsand]
  (10.38,1.78) -- (10.82,1.56) -- (10.82,2.14) -- (10.38,2.36) -- cycle;
\filldraw[iconline,fill=workflowlavender]
  (10.82,1.56) -- (11.26,1.78) -- (11.26,2.36) -- (10.82,2.14) -- cycle;
\draw[workflowink,line width=.6pt] (10.52,1.82) -- (10.99,2.06);
\fill[reportpurple] (10.94,1.80) circle (.04cm);

\filldraw[iconline,fill=workflowgreen]
  (12.92,1.66) -- (14.00,1.13) -- (15.08,1.66) -- (14.00,2.19) -- cycle;
\fill[workflowblue]
  (12.98,1.64) .. controls (13.35,1.43) and (13.55,1.69) .. (13.78,1.54)
  -- (14.03,1.66) .. controls (13.63,1.95) and (13.33,1.61) .. (12.98,1.81) -- cycle;
\fill[workflowslate]
  (13.02,1.48) -- (13.23,1.38) -- (14.94,2.02) -- (14.74,2.12) -- cycle;
\fill[workflowslate]
  (13.77,2.02) -- (13.98,1.92) -- (14.53,1.26) -- (14.32,1.36) -- cycle;
\filldraw[iconline,fill=workflowsand]
  (13.55,1.83) -- (13.84,1.69) -- (13.84,2.26) -- (13.55,2.40) -- cycle;
\filldraw[iconline,fill=workflowlavender]
  (13.84,1.69) -- (14.10,1.82) -- (14.10,2.39) -- (13.84,2.26) -- cycle;
\filldraw[iconline,fill=white]
  (13.55,2.40) -- (13.84,2.26) -- (14.10,2.39) -- (13.82,2.54) -- cycle;
\draw[fine] (13.68,1.96) -- (13.68,2.23);
\draw[fine] (13.96,1.89) -- (13.96,2.16);
\filldraw[iconline,fill=workflowlavender]
  (14.38,1.59) -- (14.58,1.49) -- (14.58,1.83) -- (14.38,1.93) -- cycle;
\filldraw[iconline,fill=workflowsand]
  (14.58,1.49) -- (14.77,1.59) -- (14.77,1.93) -- (14.58,1.83) -- cycle;
\filldraw[iconline,fill=white]
  (14.38,1.93) -- (14.58,1.83) -- (14.77,1.93) -- (14.57,2.03) -- cycle;
\draw[workflowink,line width=.65pt] (14.28,1.77) -- (14.28,2.13);
\filldraw[iconline,fill=workflowgreen!75!reportpurple]
  (14.28,2.29) -- (14.08,2.02) -- (14.48,2.02) -- cycle;
\filldraw[iconline,fill=workflowblue]
  (13.10,1.67) -- (13.30,1.57) -- (13.30,1.90) -- (13.10,2.00) -- cycle;
\filldraw[iconline,fill=workflowsand]
  (13.30,1.57) -- (13.48,1.66) -- (13.48,1.99) -- (13.30,1.90) -- cycle;
\filldraw[iconline,fill=white]
  (13.10,2.00) -- (13.30,1.90) -- (13.48,1.99) -- (13.29,2.09) -- cycle;
\draw[workflowink,line width=.55pt] (14.82,1.70) -- (14.82,1.98);
\filldraw[iconline,fill=workflowdarkgreen]
  (14.82,2.12) -- (14.66,1.92) -- (14.98,1.92) -- cycle;
\fill[reportpurple] (13.35,1.40) circle (.04cm);
\fill[reportpurple] (14.61,1.75) circle (.04cm);

\draw[flow] (3.15,1.96) -- node[edgelabel,above]{condition} (4.47,1.96);
\draw[flow] (7.45,1.96) -- node[edgelabel,above]{accepted plan} (8.36,1.96);
\draw[flow] (11.42,1.96) -- node[edgelabel,above]{compile} (12.68,1.96);

\node[note,text=workflowblue!55!workflowink] at (2.00,.62)
  {geography + assets + rules};
\node[note,text=reportpurple] at (6.00,.62)
  {regional intent + tile semantics};
\node[note,text=workflowcoral] at (10.00,.62)
  {deterministic code + constraints};
\node[note,text=workflowdarkgreen] at (14.00,.62)
  {playable geometry + state};

\end{tikzpicture}}
\caption{WorldStream workflow. World knowledge conditions a hierarchical Macro--Micro planner; local code compiles the accepted plan into a streamed, interactive 3D world. Plans, identities, seeds, and edits persist independently of rendered geometry, enabling consistent interaction and revisit.}
\label{fig:workflow-overview}
\end{figure}
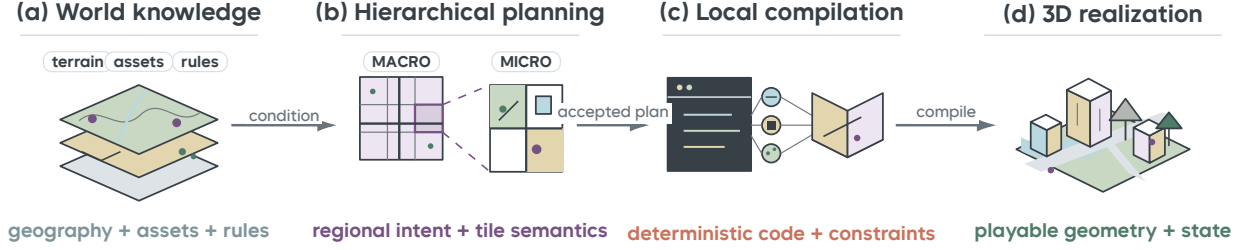

Persistent execution builds on this specification by assigning each place a retained tile record. That record combines the accepted plan with boundary commitments, stable object state, revisions, and user edits; loaded geometry is a reproducible realization of this retained state. Eviction releases only the realization, and revisit recompiles the accepted place under a bounded render-memory budget. Language interaction updates the same record through entity actions, current-region edits, and future-region intent, and each accepted change becomes a revision consumed by later planning, construction, and reconstruction. Persistent identity, streaming residency, and grounded interaction therefore operate through one shared state model. Figure~\ref{fig:workflow-overview} summarizes this organization; Section~\ref{sec:evaluation} isolates each mechanism and verifies its effect.

\begin{samepage}
Our contributions are:
\begin{itemize}
\item \textbf{Dual-frontier streaming under bounded residency,} with a semantic frontier planning connected regions ahead of movement and a narrower realization frontier compiling only the geometry needed in the near future. Direction-aware admission bounds each stage without retaining the full explored world.
\item \textbf{Geography-conditioned hierarchical planning with executable spatial contracts,} where Macro decisions organize multi-tile structure and Micro decisions specialize it under measured terrain and accepted boundaries. Joint resolution converts typed probability distributions into feasible plans before compilation.
\item \textbf{Revisioned world state for durable interaction,} retaining plans, seeds, boundary commitments, stable identities, mutable actor state, and user edits beyond geometry. It grounds current and future actions, rejects stale asynchronous updates, and drives reconstruction.
\end{itemize}
\end{samepage}

\section{Persistent World Representation for Streaming and Revisits}
\label{sec:representation}

Streaming changes resident geometry from moment to moment. \system{} represents the world with persistent tile records and derives the current rendered view from them. The record definition and reconstruction conditions establish stable identity across eviction and revisit.

To make this separation operational, \system{} uses the tile as the common unit of persistent state and runtime work. Micro planning, local compilation, scheduling, loading, and eviction all act on fixed-size tiles. Each tile corresponds to one identifiable locale---several city blocks, a village with its surrounding fields, or a section of forest or coast---while its dimensions remain a runtime configuration. Its record retains the accepted plan, source geographic evidence, boundary commitments, entity state, and user edits. Planning thereby receives a stable spatial substrate, interaction receives addressable targets, and rendering derives its resources from the same authoritative record.

Formally, the retained record is
\begin{equation}
R_i=(G_i,M_i,B_i,O_i,U_i,\rho_i,V_i),
\label{eq:tile-record}
\end{equation}
where $G_i$ is geographic evidence, $M_i$ the accepted semantic plan, $B_i$ the shared boundary commitments, $O_i$ explicit object state, $U_i$ user edits and future intent, $\rho_i$ the monotonically increasing revision, and $V_i$ the compatible generator and asset versions. The loaded realization is derived from this record,
\begin{equation}
X_i=\operatorname{Compile}(R_i).
\label{eq:record-compile}
\end{equation}
The runtime treats $R_i$ as authoritative: eviction removes $X_i$, whereas planning and editing create a new revision of $R_i$. This relationship also defines the tile lifecycle:

\begingroup
\begin{enumerate}
\setlength{\itemsep}{0.08em}
\setlength{\parskip}{0pt}
\setlength{\parsep}{0pt}
\setlength{\topsep}{0.35em}
\item \textbf{Sample geographic evidence.} Read stable height, flatness, and water values from the map or seeded surface (Section~\ref{sec:evidence}).
\item \textbf{Plan semantic content.} Macro establishes regional structure, while Micro resolves executable local content (Section~\ref{sec:planning}).
\item \textbf{Compile playable geometry.} Local generators build the planned terrain, roads, buildings, vegetation, and entities (Section~\ref{sec:realization}).
\item \textbf{Retain world state.} Plans, identities, seeds, revisions, and accepted edits survive beyond the active render cache (Section~\ref{sec:residency}).
\item \textbf{Reconstruct on revisit.} Returning tiles deterministically rebuild the accepted place from those retained records (Section~\ref{sec:consistency}).
\end{enumerate}
\endgroup

\subsection{Geographic Evidence and the Tile Substrate}
\label{sec:evidence}

Geographic evidence provides the common physical substrate for planning and realization. Each tile carries coordinate-aligned height, slope, water, and surface samples over its footprint, drawn from an existing map or a deterministic seeded surface, following established terrain representations based on procedural and hydrological structure~\cite{musgrave1993landscape,genevaux2013terrain}. These values constrain semantic decisions and later drive terrain construction, collision, placement, and navigation. Sampling remains stable in world coordinates, so neighboring tiles agree at their shared boundary while being processed independently. Tile boundaries therefore divide work while preserving surface continuity. Adjacent tiles form a \emph{Macro region} spanning watersheds, buildable corridors, and ecological transitions.

\subsection{Retained State and Reconstruction}

Once a plan has been accepted, realization completes the tile's persistent identity by fixing stable seeds, explicit object state, and compatible generator and asset versions in $R_i$. The displayed geometry is one compiled realization of that identity. Reconstruction recompiles the same record, while an edit creates a new revision consumed by later compilations. Because planning, compilation, and editing commit to records, these operations can proceed concurrently across tiles. Section~\ref{sec:interaction} details revision commit, and Section~\ref{sec:consistency} gives the conditions for faithful recompilation.

\section{Geography-Aware Hierarchical World Planning}
\label{sec:planning}

Each new place connects to accepted neighbors, conforms to the measured surface, and yields an executable specification. \system{} distributes this decision across two spatial scales, resolves their output jointly, and realizes the accepted specification with deterministic local code.

\subsection{From Regional Structure to Tile Plans}

At the regional scale, Macro planning establishes the context inherited by every tile. It operates on relief and flatness distributions, drainage and shoreline extent, and the land-use levels and road ports accepted along planned boundaries. From these inputs it determines settlement gradients, ecological structure, transport corridors, and water organization across the region.

\begin{figure}[!t]
\centering
\includegraphics[width=\textwidth]{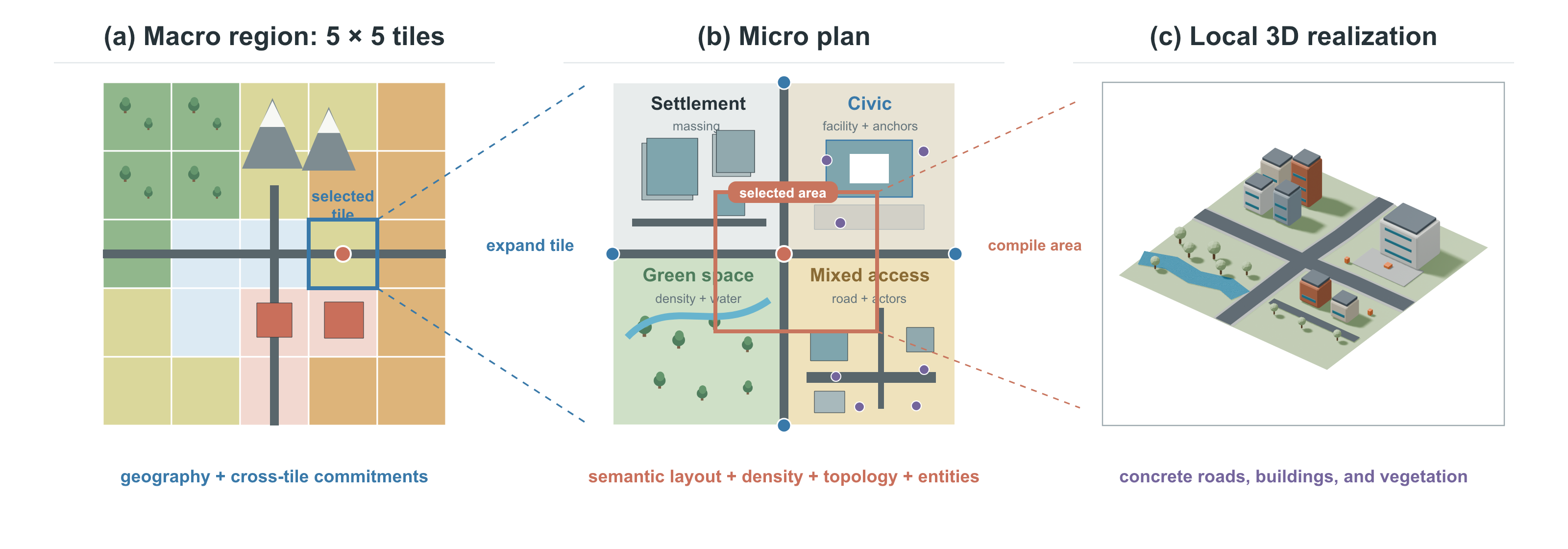}
\caption{WorldStream planning and realization. (a) Macro planning combines geographic evidence with cross-tile commitments. (b) A selected tile expands into a structured Micro plan encoding land use, roads, facilities, vegetation, water setbacks, and entity anchors. (c) Local generators realize that plan as connected roads, varied buildings, civic space, and a planted waterfront.}
\label{fig:hierarchical-planning}
\end{figure}

The regional context becomes an explicit input to Micro planning. When the scheduler admits tile $i$, the Micro Planner combines finer surface evidence with the covering Macro contract and the accepted commitments of adjacent tiles. It then derives tile-level semantics and generator parameters. Let $G_r$ and $g_i$ denote regional and tile-level geographic evidence, and let $N_r$ and $N_i$ denote accepted nearby state at the two scales. The two-stage data flow is
\begin{equation}
\begin{aligned}
M_r &= P_M(G_r,N_r),\\
m_i &= P_\mu(g_i,M_r,N_i).
\end{aligned}
\label{eq:planning-contracts}
\end{equation}
Here $M_r$ is the accepted regional contract and $m_i$ is the plan for tile $i$. Their schemas reflect the different spatial responsibilities: $M_r$ records regional roles, gradients, corridors, and boundary obligations, while the Micro result is
\begin{equation}
m_i=(Z_i,C_i,A_i,\Theta_i),
\label{eq:micro-contract}
\end{equation}
where $Z_i$ assigns semantic roles to local areas, $C_i$ references boundary connections, $A_i$ identifies facilities and entity anchors, and $\Theta_i$ contains generator parameters. Together these fields form the explicit interface consumed by local realization, which derives meshes and per-instance placements during compilation. Geographic knowledge and descriptions of available generators condition both planning stages: the former supplies relevant domain constraints, and the latter maps accepted decisions to supported realization procedures. Figure~\ref{fig:hierarchical-planning} follows this complete contract from regional structure to tile plan and local realization.

Applied to a specific tile, this contract is specialized according to local evidence. A settlement contract relates a center, lower-density outskirts, civic areas, and rural land along a transport corridor; its tile plans place those roles on buildable terrain and connect them to accepted road ports. An ecological contract relates forest, wetland, and transition areas to relief and water; its tile plans use moisture, elevation, slope, and distance to water to select plant communities, canopy density, clearings, and riparian vegetation. In both cases, the Macro contract defines the regional relation and the Micro plan supplies the executable local composition.

\subsection{Planner Decisions and Local Validation}

Within each tile, the hierarchical context is expressed as a coordinated semantic program. It specifies the roles and relationships of local areas together with terrain form, road structure, vegetation organization, abundance, and the generators that realize them. For every typed choice, the planner returns a preferred value and a probability distribution over allowed alternatives. Preferred values express the intended plan, while the distributions retain plausible substitutes for ambiguous terrain or boundaries and provide the scores used by joint resolution.

Because these fields are spatially coupled, their scores are resolved jointly into an executable assignment. Dense housing occupies buildable ground, water setbacks follow the measured shoreline, and land-use levels agree with the Macro role and accepted neighbors. Independent maximum-probability selection violates one or more of these constraints for 63\% of the distributions measured in Section~\ref{sec:evaluation}. The resolver removes choices excluded by the measured surface, Macro contract, or boundary commitments and evaluates the remaining combinations together. If $p_q(z_q)$ is the planner probability of assigning choice $z_q$ to local area $q$, $\mathcal{F}$ is the feasible assignment set, and $R(z,N_i)$ is a deterministic consistency term, the accepted assignment is
\begin{equation}
z^\star=\arg\max_{z\in\mathcal{F}}
\left[\sum_q\log p_q(z_q)+R(z,N_i)\right].
\label{eq:joint-resolution}
\end{equation}
The objective preserves semantic preference through the probability term and favors smooth internal and cross-tile transitions through $R$. Membership in $\mathcal{F}$ establishes executability. The accepted assignment is stored in $M_i$ and becomes the stable input to local realization and reconstruction.

\subsection{Terrain and the Asset Interface}
\label{sec:realization}

\begin{figure}[!t]
\centering
\setlength{\tabcolsep}{2pt}
\begin{tabular}{@{}ccc@{}}
{\small (a) City}&{\small (b) Rural land}&{\small (c) Forest}\\[-1mm]
\includegraphics[width=.285\textwidth]{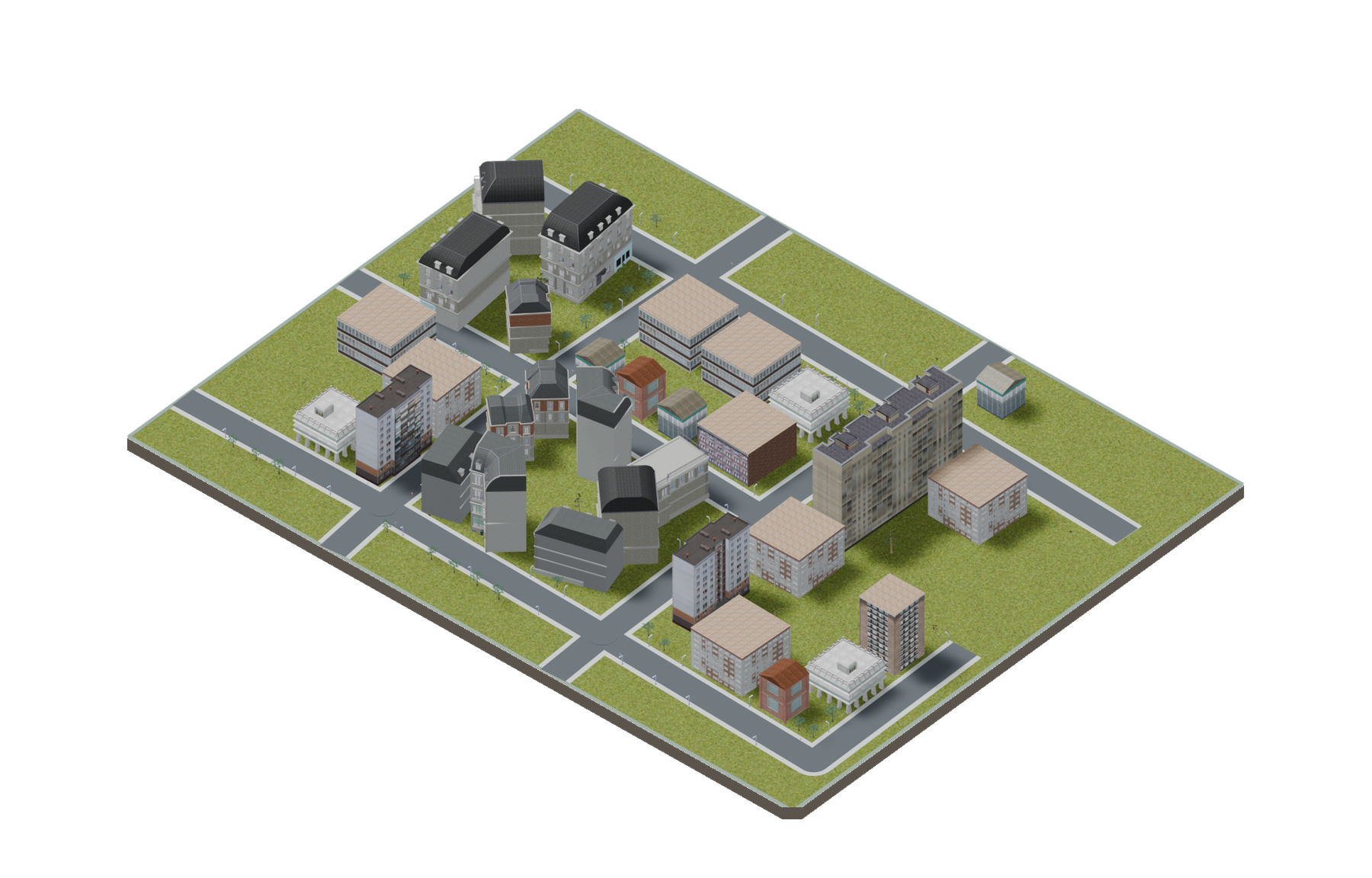}&
\includegraphics[width=.285\textwidth]{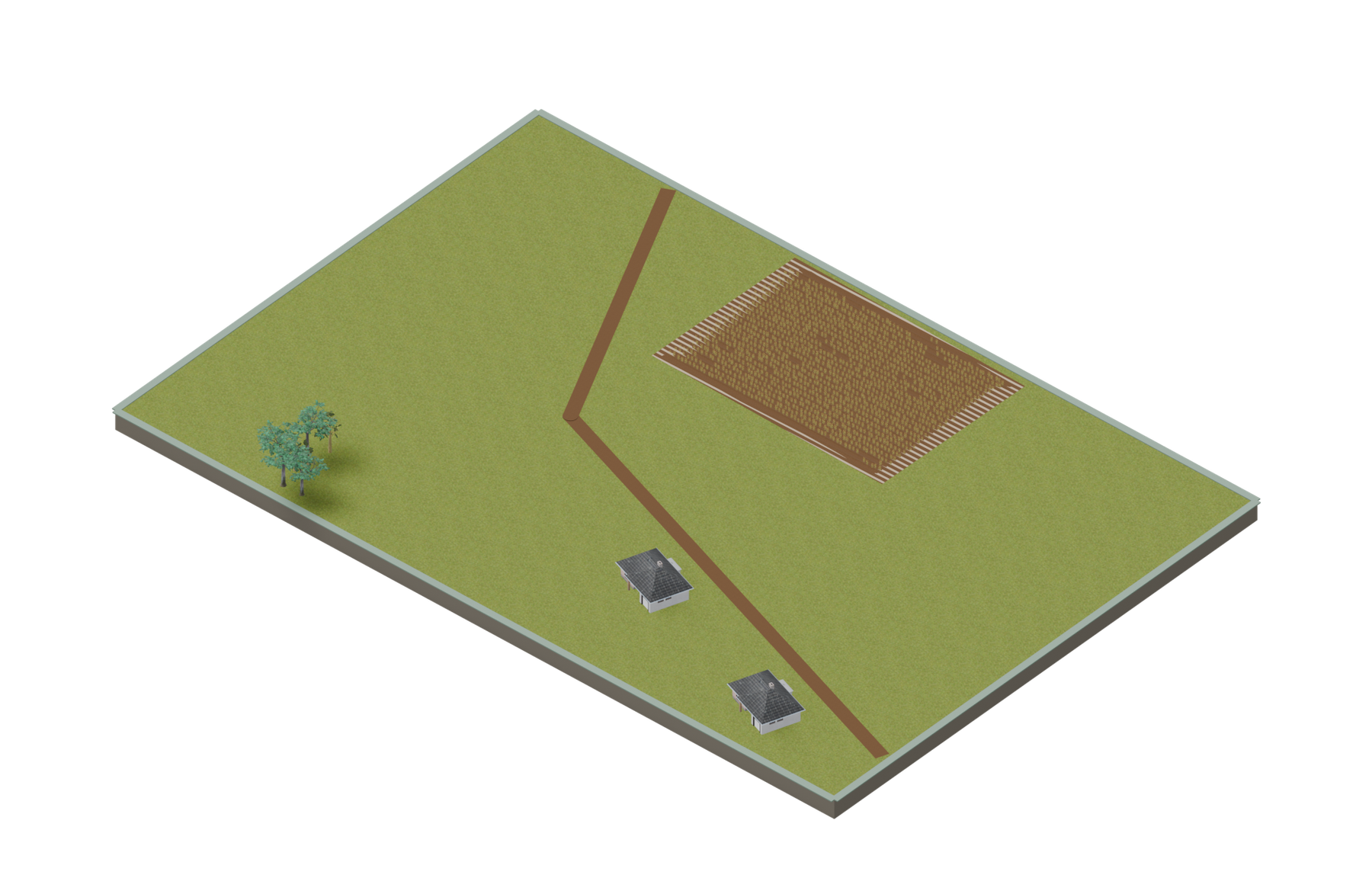}&
\includegraphics[width=.285\textwidth]{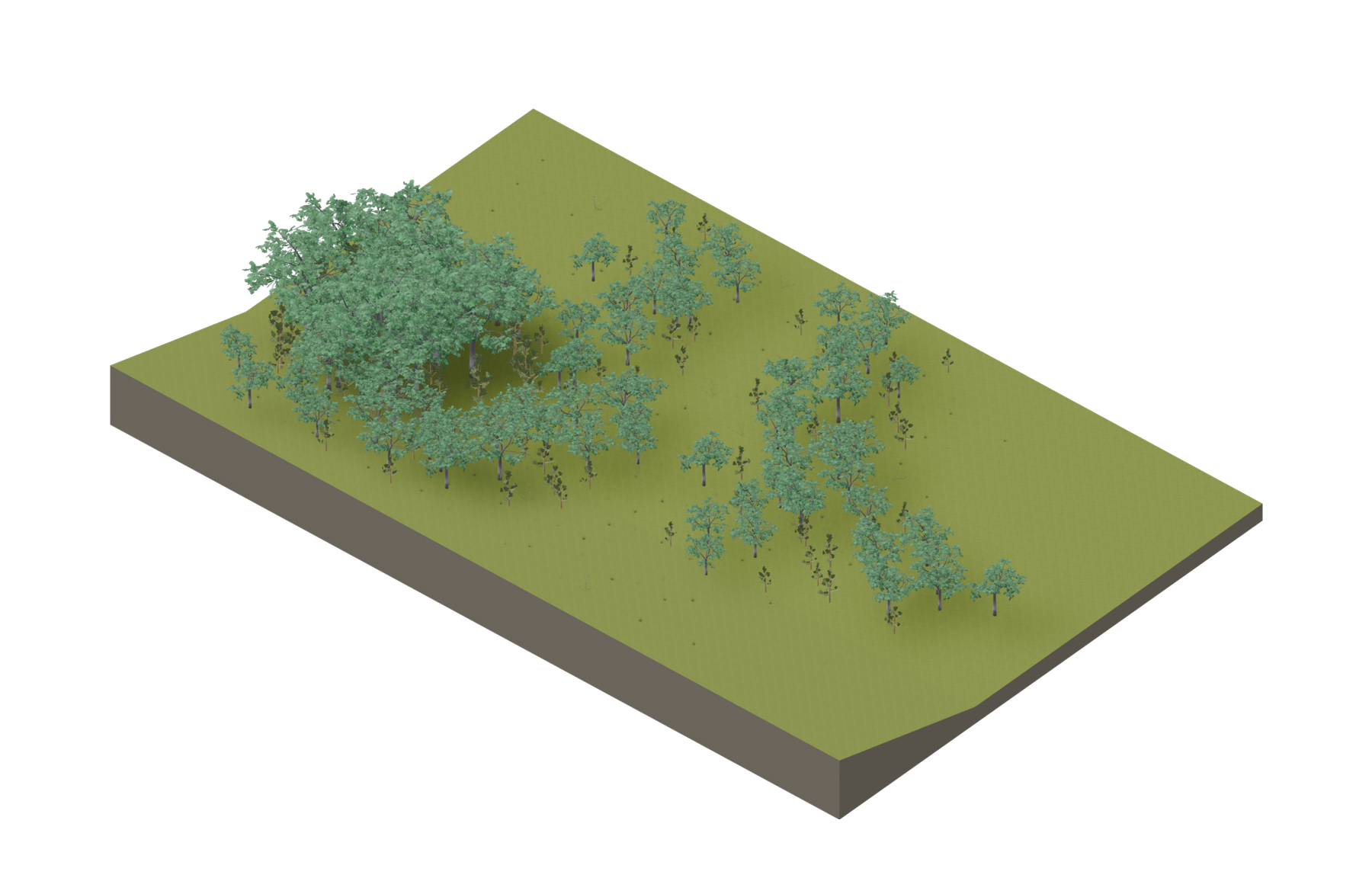}\\[1mm]
{\small (d) Mountain}&{\small (e) Coast}&{\small (f) Desert}\\[-1mm]
\includegraphics[width=.285\textwidth]{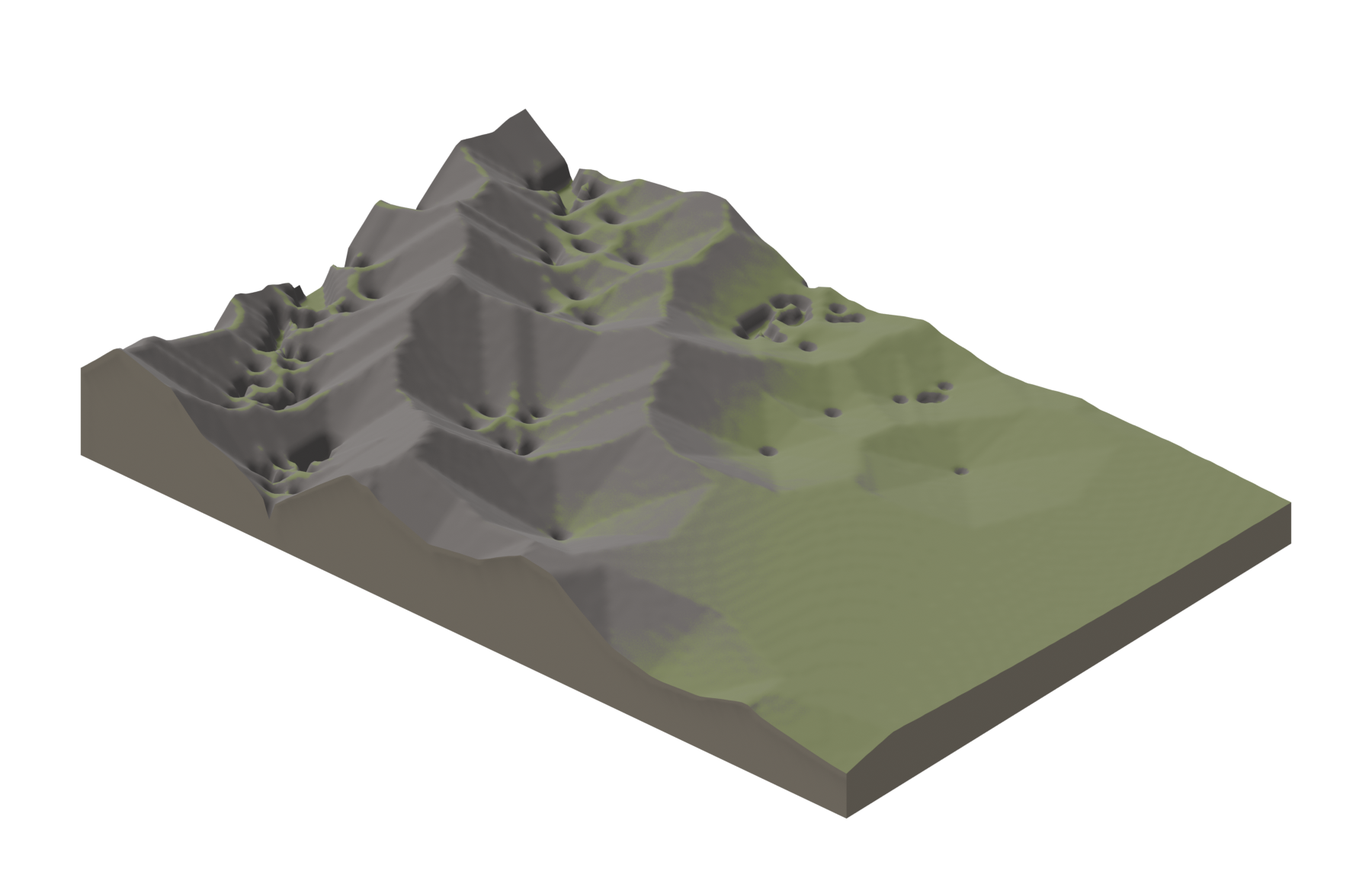}&
\includegraphics[width=.285\textwidth]{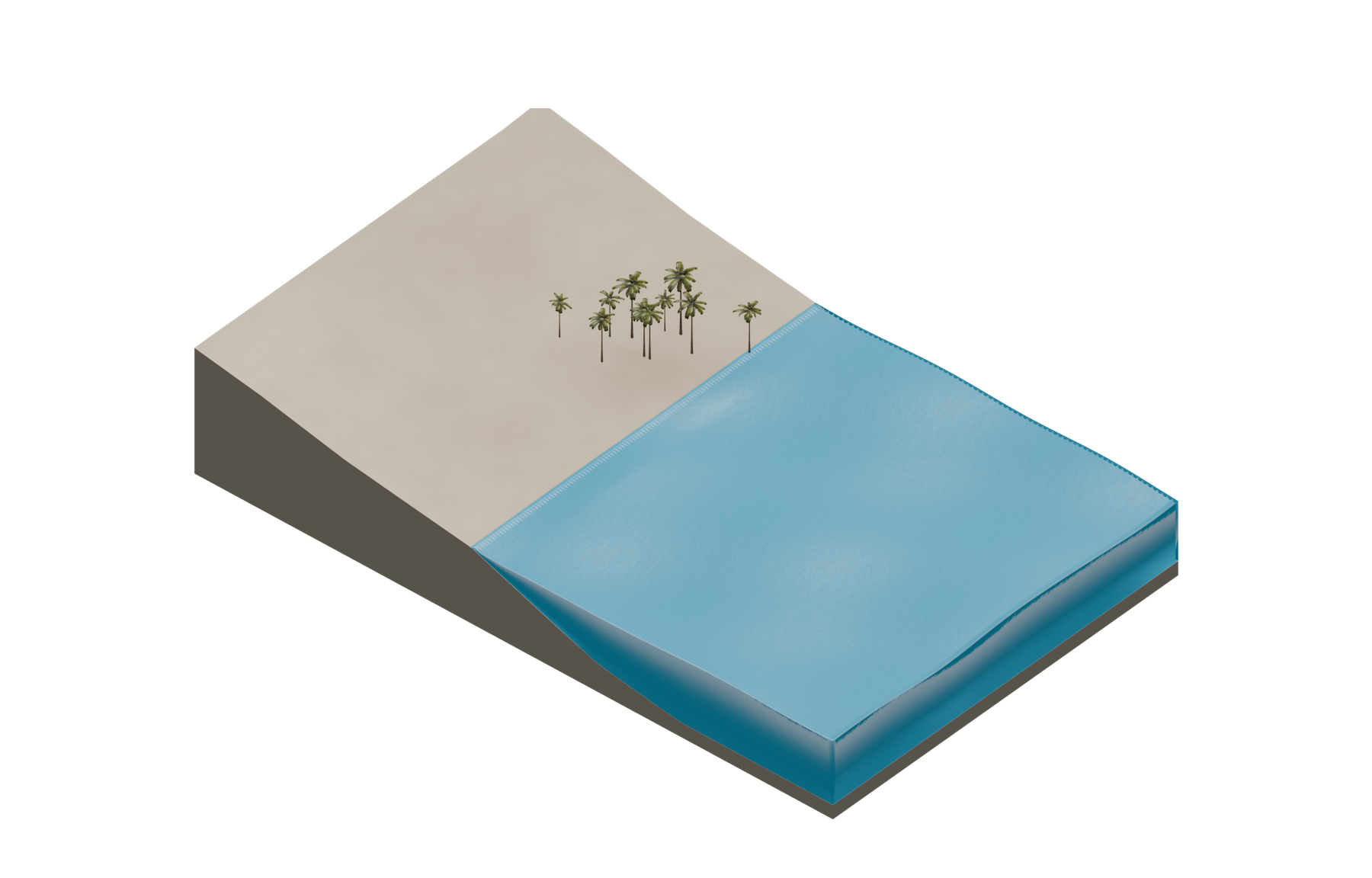}&
\includegraphics[width=.285\textwidth]{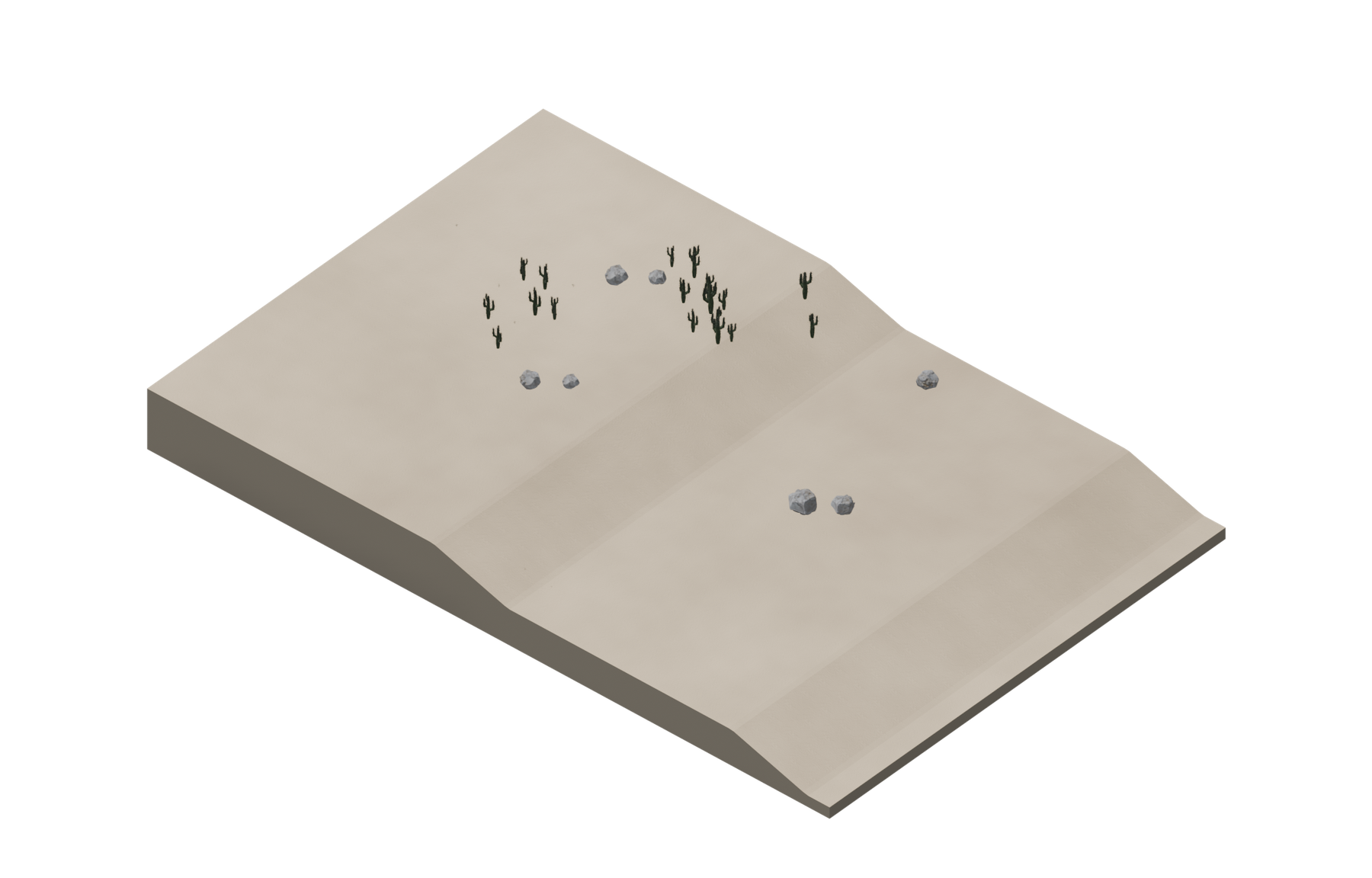}
\end{tabular}
\caption{Terrain-and-asset realizations from persisted generated worlds. Panels show the same local-generation interface across constructed and natural settings, preserving accepted plans and geographic inputs.}
\label{fig:local-families}
\end{figure}

Once $m_i$ has been accepted, deterministic local generators reproduce its geometry under a compatible implementation, establishing the accepted plan as the durable form of that place. Terrain code reconstructs height and water as the continuous geographic substrate; content-specific algorithms then populate that surface from asset-library records exposing semantic role, environmental compatibility, bounds, scale, and rendering metadata. The accepted plan and measured surface jointly determine selection and placement through a shared realization interface for roads, settlements, vegetation, and other content families. Figure~\ref{fig:local-families} shows six representative families.

Compilation follows their spatial dependencies. Terrain and water define buildable ground; road connections reserve circulation space; buildings and major objects occupy compatible footprints; vegetation and smaller details fill remaining eligible space under masks and spacing constraints.

\subsection{Content-Specific Placement}

From this shared interface, realization branches into placement algorithms matched to each content structure. Streets and building footprints follow connectivity from committed boundary ports; plant positions follow spatial correlation in a coherent field.

\paragraph{Roads and settlements.}
Road generation begins from committed boundary ports and the selected street framework, connects planned entries, and reserves sufficient space for roads and adjacent development. The regional contract supplies corridor orientation and settlement intensity, while local terrain checks define admissible ground. Building placement then uses the resulting circulation structure and available development footprints. Candidate footprints respect water, slope, occupancy, and asset-scale constraints. Settlement type and density jointly control frontage occupancy, building mix, height, and open space. This decomposition follows procedural city, example-driven street, and hierarchical terrain-aware road modeling~\cite{parish2001cities,chen2008streets,galin2010roads,galin2011hierarchicalroads,vanegas2012parcels}.

\paragraph{Natural environments.}
Natural areas require spatial correlation across their local extent. Recognizable woodland contains dense groups, gaps, edge growth, and understory, none of which survives independent uniform sampling alone. The vegetation code therefore combines seeded spatial proposals with coherent environmental fields, and filters those proposals through community-specific acceptance rules and final geometric checks~\cite{deussen1998ecosystems,newlands2022forests}. The accepted plan supplies a community, species mix, structure, and abundance; from these fields the code selects natural assets and evaluates each candidate against biome compatibility, local slope and relief, land--water masks, and proximity to actual water. Accepted trees and rocks are further subject to trunk and crown spacing checks, after which shrubs and smaller details populate the remaining compatible ground.

Three quantities drive the vegetation compiler at ground position $x$: an abundance parameter $\alpha$, a coherent field $\phi(x)\in[0,1]$, and local suitability $s(x)\in[0,1]$. A community profile $h_c$ turns them into the proposal probability
\begin{equation}
\pi_c(x)=\operatorname{clamp}\!\left(\alpha\,s(x)\,h_c(\phi(x)),0,1\right).
\end{equation}
Community profiles map the coherent field to different spatial structures; masks, spacing, eligibility, and budgets filter the proposals. Stable seeds make the result repeatable, while minimum-distance sampling provides a related spacing mechanism~\cite{bridson2007poisson}.

\section{Streaming Runtime and Persistent Residency}
\label{sec:streaming}

Continuous traversal produces a demand boundary that moves with the player. The runtime anticipates upcoming tiles, admits planning and construction under finite budgets, and releases geometry behind the player while preserving retained state. Figure~\ref{fig:scheduling} summarizes the progression from the active neighborhood to direction-aware prefetch and budgeted admission.

\subsection{Movement-Driven Demand and Prefetch}
\label{sec:demand-sets}

To convert motion into spatial demand, the scheduler assigns unseen tiles to three roles. The \emph{required} set $\mathcal{R}$ covers the player's immediate neighborhood and receives the strongest priority. A larger \emph{preferred} set $\mathcal{P}$ maintains the visible and near-future surroundings. Beyond it, directional prefetch supplies speculative candidates along movement and camera direction. These windows move over the tile lattice independently of the fixed spatial support used by Macro planning.

Write $\mathcal{P}(\mathbf{c},b)$ for the preferred window of half-extent $b$ centered at world position $\mathbf{c}$. Given velocity $\mathbf{v}$ and planning horizon $H_p$, the motion-predicted candidate set is the sweep of that window along the predicted path,
\begin{equation}
\mathcal{A}(H_p)=\bigcup_{0\leq\tau\leq H_p}\mathcal{P}(\mathbf{p}+\tau\mathbf{v},b),\qquad
\eta(i)=\inf\{\tau:i\in\mathcal{P}(\mathbf{p}+\tau\mathbf{v},b)\}.
\end{equation}
Here $\mathbf{p}$ and $\mathbf{v}$ are the current position and velocity, and $\eta(i)$ is the predicted time at which tile $i$ first enters the moving preferred window. The implementation evaluates window-crossing events, including diagonal crossings. A camera-facing supplement $\mathcal{V}$ prepares content when view direction carries useful information or the player is stationary; the full speculative candidate set is $\mathcal{G}=\mathcal{A}\cup\mathcal{V}$.

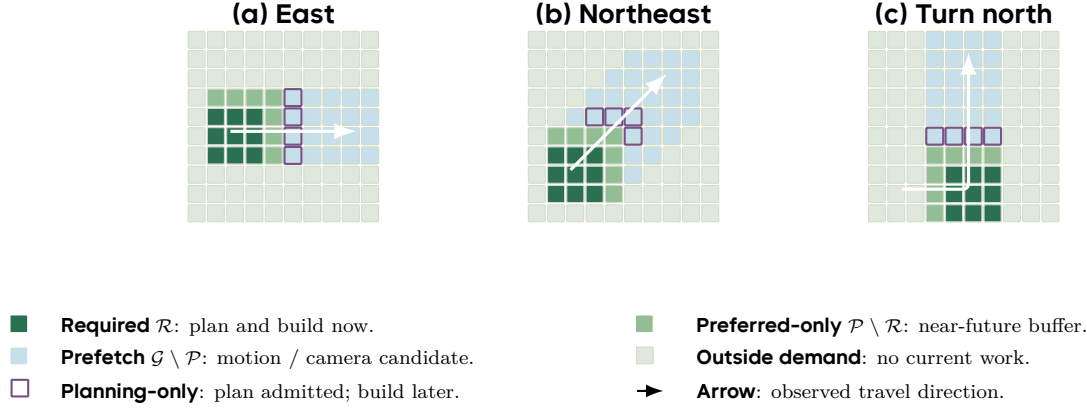
\begin{figure}[!t]
\centering
\definecolor{scheduleRequired}{HTML}{2B7050}
\definecolor{schedulePreferred}{HTML}{93BD94}
\definecolor{schedulePrefetch}{HTML}{C7DFEA}
\definecolor{scheduleInactive}{HTML}{E0E7DC}
\definecolor{scheduleBorder}{HTML}{D0DBC8}
\begin{tikzpicture}[x=0.255cm,y=-0.255cm,font=\footnotesize\sffamily]
\foreach \shift/\title in {0/{(a) East},4.5/{(b) Northeast},9/{(c) Turn north}}{
  \begin{scope}[xshift=\shift cm]
    \foreach \y in {0,...,9}{
      \foreach \x in {0,...,9}{
        \draw[fill=scheduleInactive,draw=scheduleBorder,line width=.25pt,rounded corners=.45pt]
          ({\x+.06},{\y+.06}) rectangle ({\x+.94},{\y+.94});
      }
    }
    \node[font=\small\sffamily\bfseries] at (5,-.85) {\title};
  \end{scope}
}

\begin{scope}
  \foreach \y in {3,...,6}{\foreach \x in {5,...,9}{
    \draw[fill=schedulePrefetch,draw=scheduleBorder,line width=.25pt,rounded corners=.45pt]
      ({\x+.06},{\y+.06}) rectangle ({\x+.94},{\y+.94});}}
  \foreach \y in {3,...,6}{\foreach \x in {1,...,4}{
    \draw[fill=schedulePreferred,draw=scheduleBorder,line width=.25pt,rounded corners=.45pt]
      ({\x+.06},{\y+.06}) rectangle ({\x+.94},{\y+.94});}}
  \foreach \y in {4,...,6}{\foreach \x in {1,...,3}{
    \draw[fill=scheduleRequired,draw=scheduleBorder,line width=.25pt,rounded corners=.45pt]
      ({\x+.06},{\y+.06}) rectangle ({\x+.94},{\y+.94});}}
  \foreach \y in {3,...,6}{\draw[youtuBlue,line width=.8pt,rounded corners=.45pt]
    (5.06,{\y+.06}) rectangle (5.94,{\y+.94});}
  \draw[-{Latex},very thick,white] (2.25,5.25)--(8.8,5.25);
\end{scope}

\begin{scope}[xshift=4.5cm]
  \foreach \x/\y in {5/1,6/1,7/1,8/1,4/2,5/2,6/2,7/2,8/2,3/3,4/3,5/3,6/3,7/3,8/3,2/4,3/4,4/4,5/4,6/4,7/4,8/4,5/5,6/5,7/5,5/6,6/6,5/7}{
    \draw[fill=schedulePrefetch,draw=scheduleBorder,line width=.25pt,rounded corners=.45pt]
      ({\x+.06},{\y+.06}) rectangle ({\x+.94},{\y+.94});}
  \foreach \y in {5,...,8}{\foreach \x in {1,...,4}{
    \draw[fill=schedulePreferred,draw=scheduleBorder,line width=.25pt,rounded corners=.45pt]
      ({\x+.06},{\y+.06}) rectangle ({\x+.94},{\y+.94});}}
  \foreach \y in {6,...,8}{\foreach \x in {1,...,3}{
    \draw[fill=scheduleRequired,draw=scheduleBorder,line width=.25pt,rounded corners=.45pt]
      ({\x+.06},{\y+.06}) rectangle ({\x+.94},{\y+.94});}}
  \foreach \x/\y in {3/4,4/4,5/4,5/5}{\draw[youtuBlue,line width=.8pt,rounded corners=.45pt]
    ({\x+.06},{\y+.06}) rectangle ({\x+.94},{\y+.94});}
  \draw[-{Latex},very thick,white] (2.25,7.25)--(7.25,2.25);
\end{scope}

\begin{scope}[xshift=9cm]
  \foreach \y in {0,...,5}{\foreach \x in {3,...,6}{
    \draw[fill=schedulePrefetch,draw=scheduleBorder,line width=.25pt,rounded corners=.45pt]
      ({\x+.06},{\y+.06}) rectangle ({\x+.94},{\y+.94});}}
  \foreach \y in {6,...,9}{\foreach \x in {3,...,6}{
    \draw[fill=schedulePreferred,draw=scheduleBorder,line width=.25pt,rounded corners=.45pt]
      ({\x+.06},{\y+.06}) rectangle ({\x+.94},{\y+.94});}}
  \foreach \y in {7,...,9}{\foreach \x in {4,...,6}{
    \draw[fill=scheduleRequired,draw=scheduleBorder,line width=.25pt,rounded corners=.45pt]
      ({\x+.06},{\y+.06}) rectangle ({\x+.94},{\y+.94});}}
  \foreach \x in {3,...,6}{\draw[youtuBlue,line width=.8pt,rounded corners=.45pt]
    ({\x+.06},5.06) rectangle ({\x+.94},5.94);}
  \draw[-{Latex},very thick,white,rounded corners=2pt] (1.8,8.25)--(5.25,8.25)--(5.25,1.15);
\end{scope}

\end{tikzpicture}
\par\vspace{12mm}
\fontsize{7.5}{8.7}\selectfont
\begin{tabularx}{.98\textwidth}{@{}lXlX@{}}
\tikz\draw[fill=scheduleRequired,draw=none,rounded corners=.3pt] (0,0) rectangle (.22,.22); &
\textbf{Required $\mathcal{R}$}: plan and build now. &
\tikz\draw[fill=schedulePreferred,draw=none,rounded corners=.3pt] (0,0) rectangle (.22,.22); &
\textbf{Preferred-only $\mathcal{P}\setminus\mathcal{R}$}: near-future buffer.\\[1mm]
\tikz\draw[fill=schedulePrefetch,draw=none,rounded corners=.3pt] (0,0) rectangle (.22,.22); &
\textbf{Prefetch $\mathcal{G}\setminus\mathcal{P}$}: motion / camera candidate. &
\tikz\draw[fill=scheduleInactive,draw=scheduleBorder,rounded corners=.3pt] (0,0) rectangle (.22,.22); &
\textbf{Outside demand}: no current work.\\[1mm]
\tikz\draw[youtuBlue,line width=.8pt,rounded corners=.3pt] (0,0) rectangle (.22,.22); &
\textbf{Planning-only}: plan admitted; build later. &
\tikz\draw[-{Latex},line width=.8pt] (0,.11)--(.35,.11); &
\textbf{Arrow}: observed travel direction.
\end{tabularx}
\caption{Directional demand and budgeted admission on a square tile lattice. Dark green tiles are required immediately; light green tiles extend the visible and near-future buffer; pale blue tiles are speculative motion- or camera-facing candidates; gray tiles lie outside current demand. A blue outline marks a speculative tile that survives the planning budget and receives a semantic plan, but remains outside the shorter construction horizon. White arrows show observed travel direction. On square tiles, northeast motion crosses the north and east boundaries together and produces an L-shaped first-arrival front; the third panel shows corridor recomputation after an observed turn.}
\label{fig:scheduling}
\end{figure}

Under this demand model, predicted entry time determines how work advances along the route. Figure~\ref{fig:scheduling} contrasts the proposed corridor with the subset admitted by a fixed budget: a tile crossed in two seconds is prepared ahead of a nearer tile behind the player, extending effective lookahead along the direction travelled. Recomputing the corridor after a turn shifts priority to the new direction, while completed plans remain available for reuse.

\subsection{Budgeted Planning and Construction}
\label{sec:admission}

Budgeted admission turns predicted demand into bounded work. Motion prediction and camera-facing windows define the candidate pool, while separate budgets select candidates for planning and construction. Let $\mathcal{L}\supseteq\mathcal{R}$ denote the required neighborhood together with a one-tile halo that absorbs prediction error at its edge. Under planning budget $K_p$, construction budget $K_c\leq K_p$, and construction horizon $H_c\leq H_p$, the admitted sets are
\begin{align}
\mathcal{Q}&=\operatorname{Top}_{K_p}(\mathcal{L}\cup\mathcal{G}),\\
\mathcal{C}&=\operatorname{Top}_{K_c}
\left(\mathcal{Q}\cap\left[\mathcal{L}\cup\{i:\eta(i)\leq H_c\}\right]\right).
\end{align}
$\operatorname{Top}_K$ retains the $K$ highest-priority candidates under the corresponding budget. The shorter construction horizon makes geometry a subset of semantic preparation, while recent dispatch-to-finish times adapt both horizons to current tile-preparation cost. The resulting sets satisfy
\begin{equation}
\mathcal{C}\subseteq\mathcal{Q},\qquad
|\mathcal{Q}|\leq K_p,\qquad
|\mathcal{C}|\leq K_c.
\label{eq:admission-properties}
\end{equation}
These properties give each stage an explicit concurrency bound and ensure that compilation consumes an accepted plan. Within those bounds, priority bands place required and preferred tiles above speculative candidates; ETA, distance, camera alignment, and a stable coordinate tie-break order tiles within each band. Entry time determines how far the budget reaches along the corridor while the near field retains priority. Deduplication then shares running work, and queue reconciliation reuses jobs already in flight when a tile re-enters the window.

\subsection{Resource Residency and Reconstruction}
\label{sec:residency}

After construction, residency is represented according to content behavior. Unique and interactive objects---actors, landmarks, and user-addressable content---carry explicit identity and mutable state. Dense procedural content retains its plan, parameters, masks, seed, and compatible generator version to recover the layout deterministically. An individual edit promotes the affected object into an explicit record, allowing one edited tree to persist while the surrounding forest remains parametric. This division bounds persistent storage by addressable and modified state while keeping every addressable object restorable. Repeated objects share prototypes and materials, prepared assets are cached across tiles, and installation binds cached assets before replacing visible content.

\subsection{Consistency Across Independently Processed Tiles}
\label{sec:consistency}

Asynchronous processing introduces a second requirement: tiles must agree independently of arrival order and travel direction. Three record-level mechanisms provide this consistency.
\begin{enumerate}
\item \textbf{A shared boundary record.} Both incident tiles read one authoritative accepted record for their common edge---the canonical road port, edge elevation, and ecological transition---and route their generated geometry to it. Independent compilation of either tile therefore produces matching roads, terrain profiles, and vegetation edges along that boundary.
\item \textbf{A schedule-independent plan key.} A planning result is keyed by tile identity, geographic evidence, Macro contract, neighboring commitments, edits, and the active capability and generator versions. Look-ahead requests, in-window requests, and revisits share the same semantic result whenever these inputs agree.
\item \textbf{A revision stamp on background work.} Every planning or compilation job carries its source revision. Installation accepts results whose source revision matches the current record, preserving the latest committed edit.
\end{enumerate}
Cross-tile context also includes soft neighbor summaries. An accepted settlement level, ecological transition, or facility near a boundary shapes the next tile while allowing a distinct land-use label. Hard records preserve physical connections; soft summaries guide semantic continuity.

These three records fix what a tile means, independently of when and in what order the runtime got to it. Recompiling the record therefore reproduces the accepted place, and the explored world grows asynchronously while identity stays stable across interaction, unloading, and revisit.

\section{Grounded Language Interaction and World Editing}
\label{sec:interaction}

Language interaction reads and writes the tile records maintained by the streaming runtime. Each change is grounded in current world state and persists through a committed record revision.

\subsection{Grounded Action Selection and Execution}

Given utterance $u$, action selection combines the current world context with the available capabilities and resolves an operation with grounded parameters. The same interface covers entity actions, current-region edits, and semantic intent for future tiles. Formally,
\begin{equation}
(a,\theta)=P_a(u,W_{\mathrm{local}},\mathcal{K}),\qquad
W'=\Phi_a(W;\theta),
\end{equation}
where $u$ is the utterance, $\mathcal{K}$ the available capabilities, $W$ the world record and $W_{\mathrm{local}}$ the compact view of it supplied to the planner, $a$ the chosen action, and $\theta$ its grounded parameters. The local function $\Phi_a$ realizing $a$ then maps $W$ to a new revision $W'$. $W_{\mathrm{local}}$ carries nearby entities, points of interest, and the capability descriptions, which grounds an utterance in targets present in the world. Execution then passes to $\Phi_a$, which validates the current state and supplies the continuous detail, leaving movement, animation, collision handling, and rendering with the controllers. Table~\ref{tab:capabilities} summarizes the division across request classes.

\begin{table}[htbp]
\centering
\caption{Language-interaction classes and the division between semantic decisions and local execution.}
\label{tab:capabilities}
\small
\renewcommand{\arraystretch}{1.12}
\setlength{\tabcolsep}{7pt}
\begin{tabularx}{\textwidth}{@{}>{\raggedright\arraybackslash}p{2.85cm}>{\raggedright\arraybackslash}p{3.75cm}Y@{}}
\toprule
\textbf{Request} & \textbf{Semantic decision} & \textbf{Runtime realization}\\
\midrule
\textbf{Object query} & \tablekey{Reference} an entity or place & Read \tablekey{state} and derive its spatial relation\\
\addlinespace[.12em]
\textbf{Entity action} & \tablekey{Target + action} with parameters & Update state; \tablekey{navigate, avoid, animate}\\
\textbf{Entity creation} & \tablekey{Type + motion intent} in a spawn context & Load the asset and start its \tablekey{controller}\\
\addlinespace[.25em]
\textbf{Future intent} & \tablekey{Desired semantics} for an \tablescope{unplanned area} & Retain intent for \tablescope{later} hierarchical planning\\
\textbf{Region edit} & \tablekey{Operation + relations} over semantic fields & Recompile \tablekey{terrain, networks, ecology, assets}\\
\bottomrule
\end{tabularx}
\end{table}

\subsection{Editing Existing and Future World State}

Region editing reuses the realization pipeline through a structured semantic program. The planner predicts target and extent, spatial composition, zone or content types, shape and orientation, and generator parameters such as vegetation structure and abundance. Joint prediction preserves containment, adjacency, and connection among these fields, and local code converts the program into terrain, water, navigation, networks, and content placement.

Whether the target already exists determines when this program is realized. Existing-place edits compile against the accepted region; future-region edits retain the same fields as inputs to later Macro and Micro planning. Both paths preserve one semantic representation and converge on the same asynchronous commit protocol. Let an interaction start from tile revision $\rho_i^s$ and produce candidate update $\Delta_i$. Its commit rule is
\begin{equation}
\operatorname{Commit}(i,\Delta_i,\rho_i^s)=
\begin{cases}
\operatorname{Apply}(R_i,\Delta_i)^{\rho_i^s+1}, & \text{if }\rho_i=\rho_i^s\text{ and the target remains valid},\\
\varnothing, & \text{otherwise}.
\end{cases}
\label{eq:revision-commit}
\end{equation}
Revision validation is the final gate between prepared work and world state. Resources are prepared first; successful validation atomically installs visible state and the new retained record. This optimistic rule~\cite{kung1981optimistic} preserves the latest revision for later interaction and reconstruction.

\section{Experimental Evaluation}
\label{sec:evaluation}

\begin{figure}[!t]
\centering
\setlength{\tabcolsep}{1.3pt}
\begin{tabular}{ccc}
\includegraphics[width=.322\textwidth]{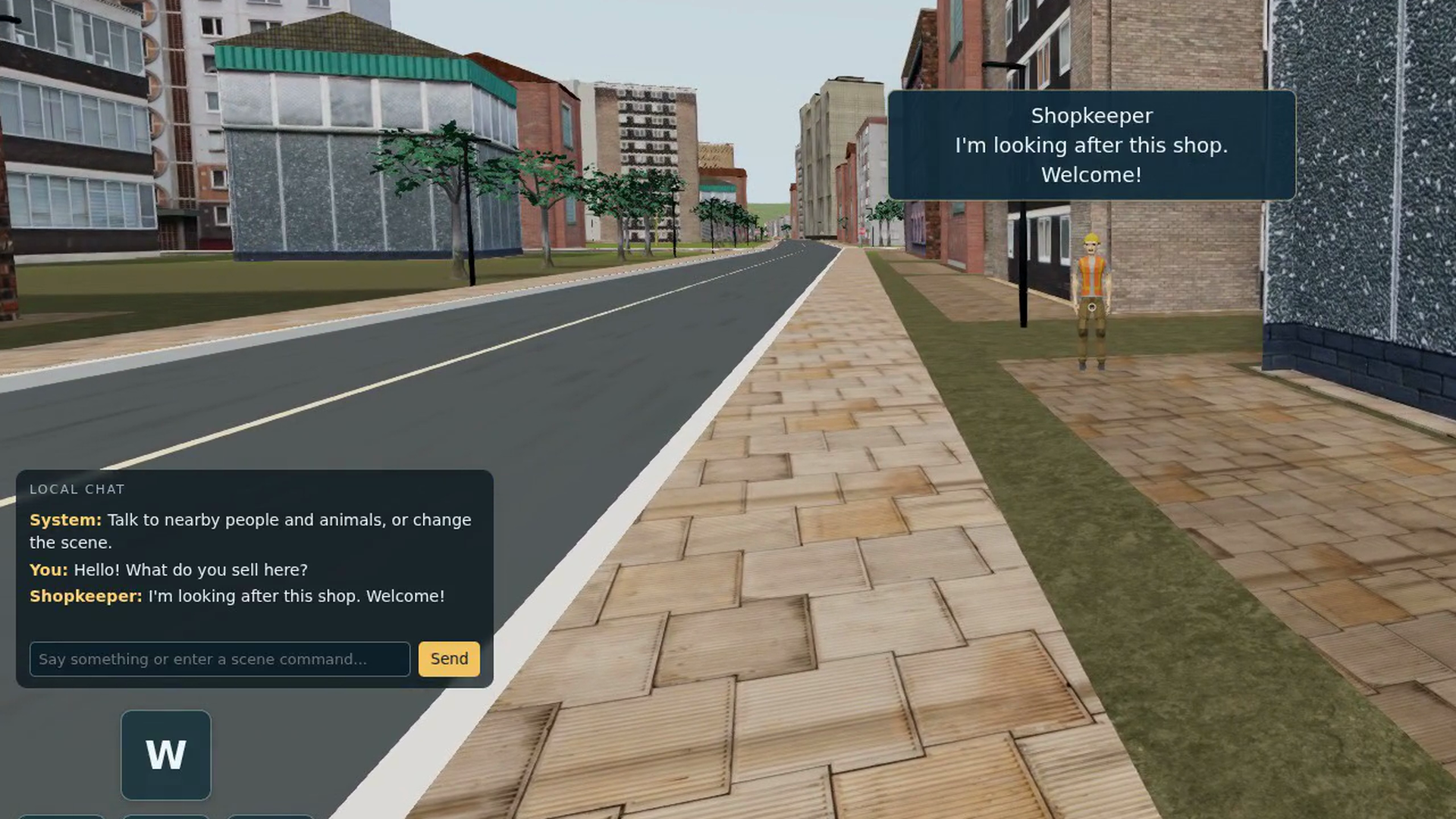}&
\includegraphics[width=.322\textwidth]{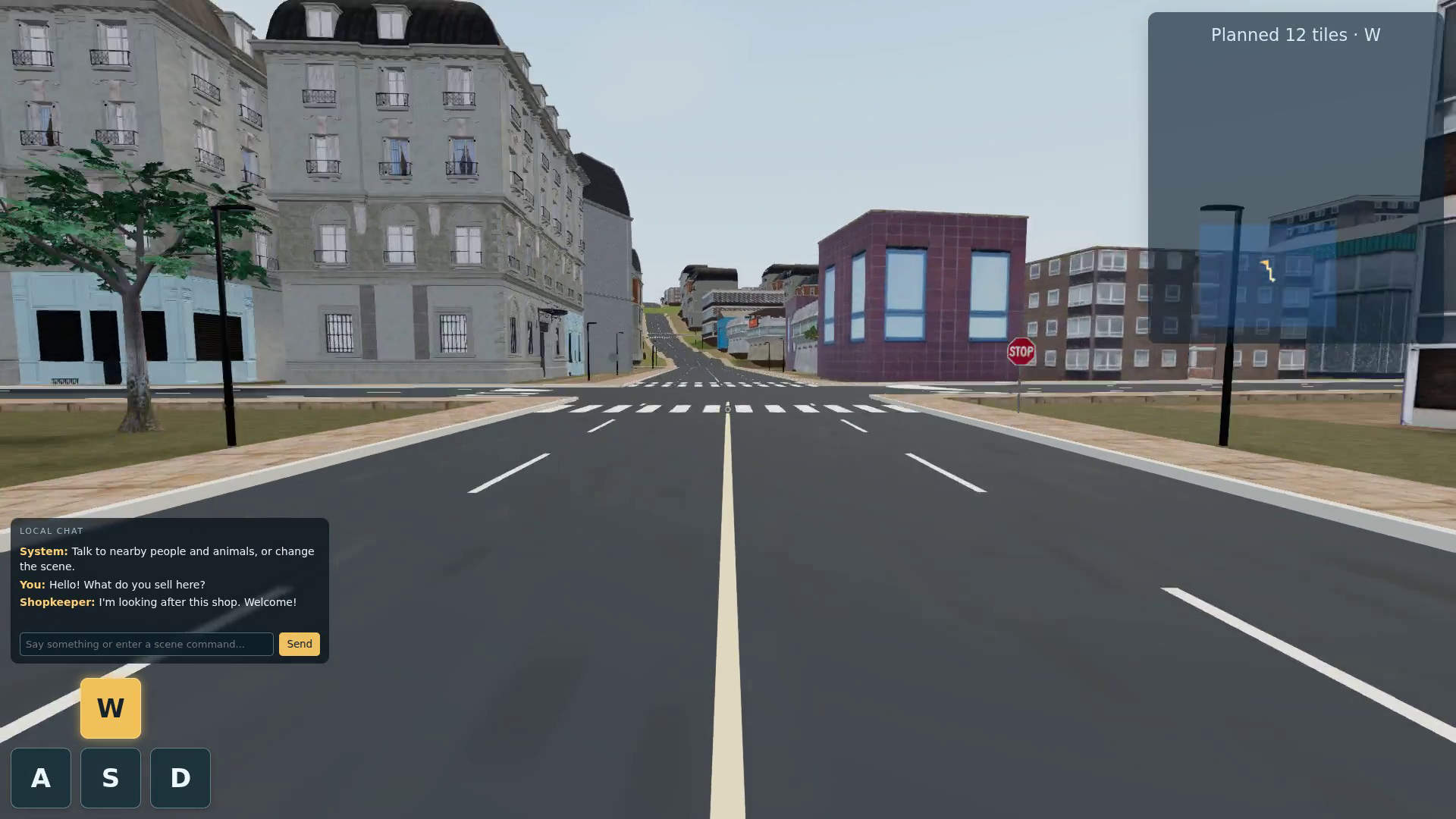}&
\includegraphics[width=.322\textwidth]{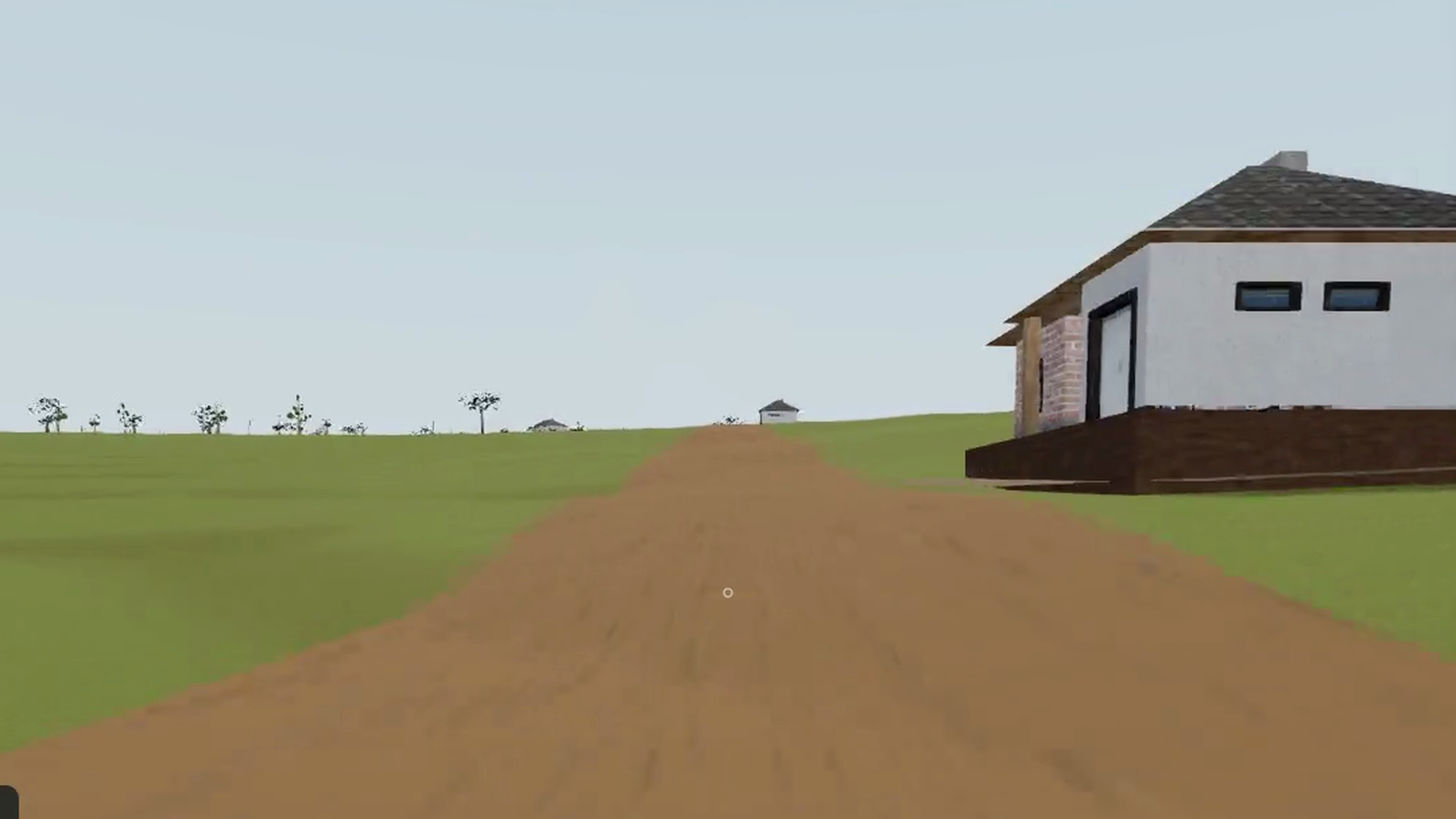}\\[-1mm]
{\small (a) City NPC}&{\small (b) City district}&{\small (c) Rural road}\\[1.5mm]
\includegraphics[width=.322\textwidth]{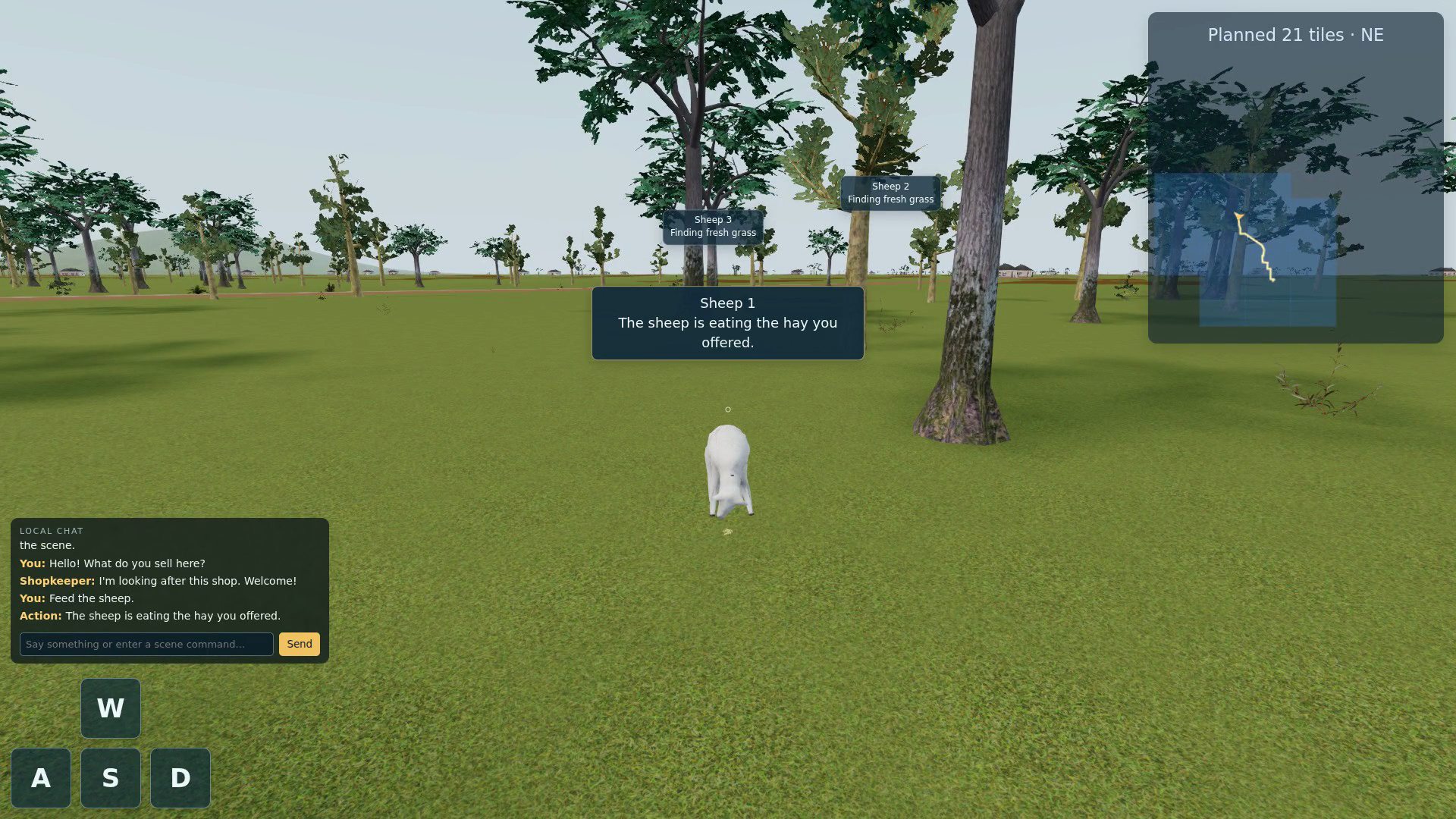}&
\includegraphics[width=.322\textwidth]{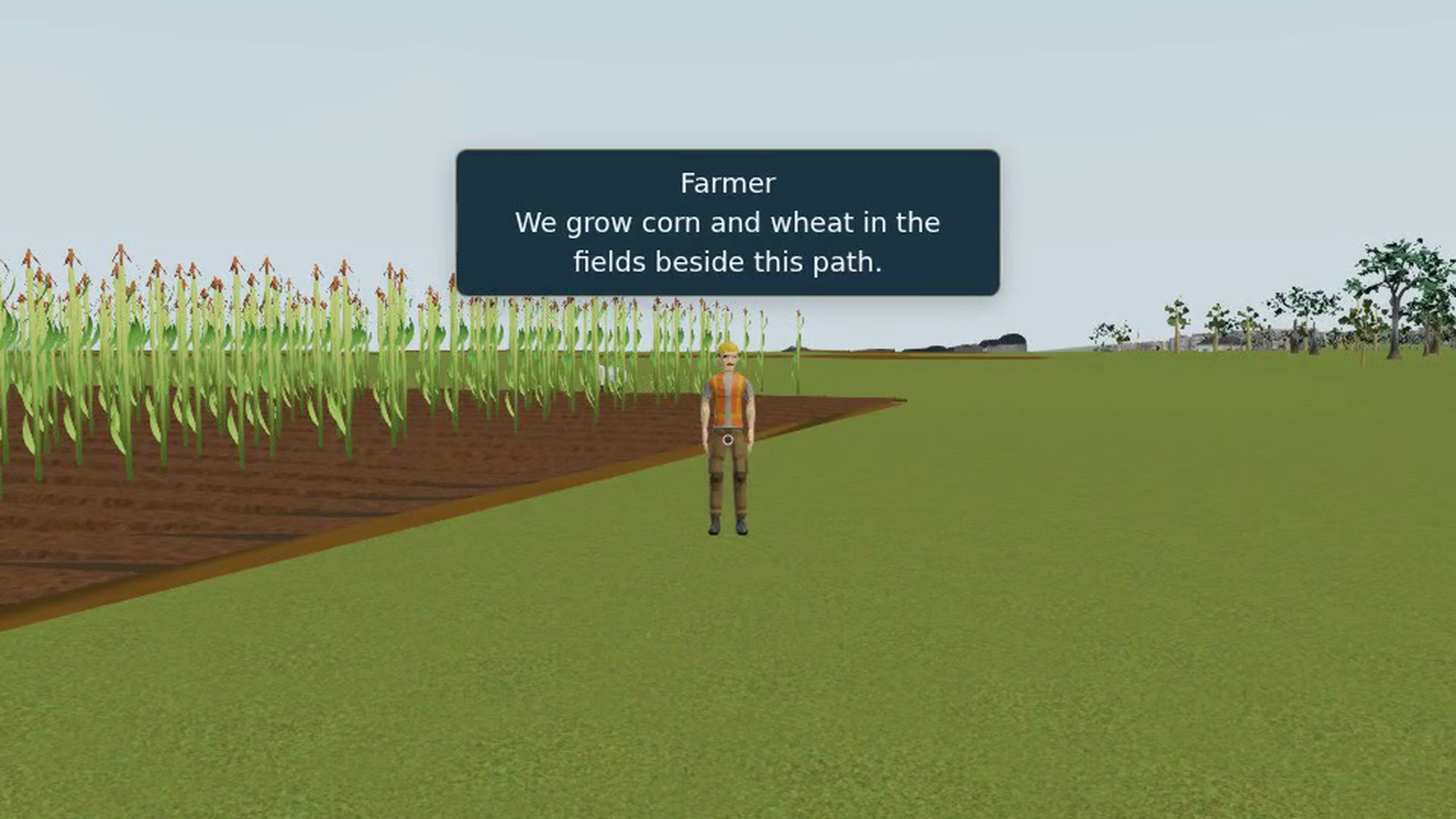}&
\includegraphics[width=.322\textwidth]{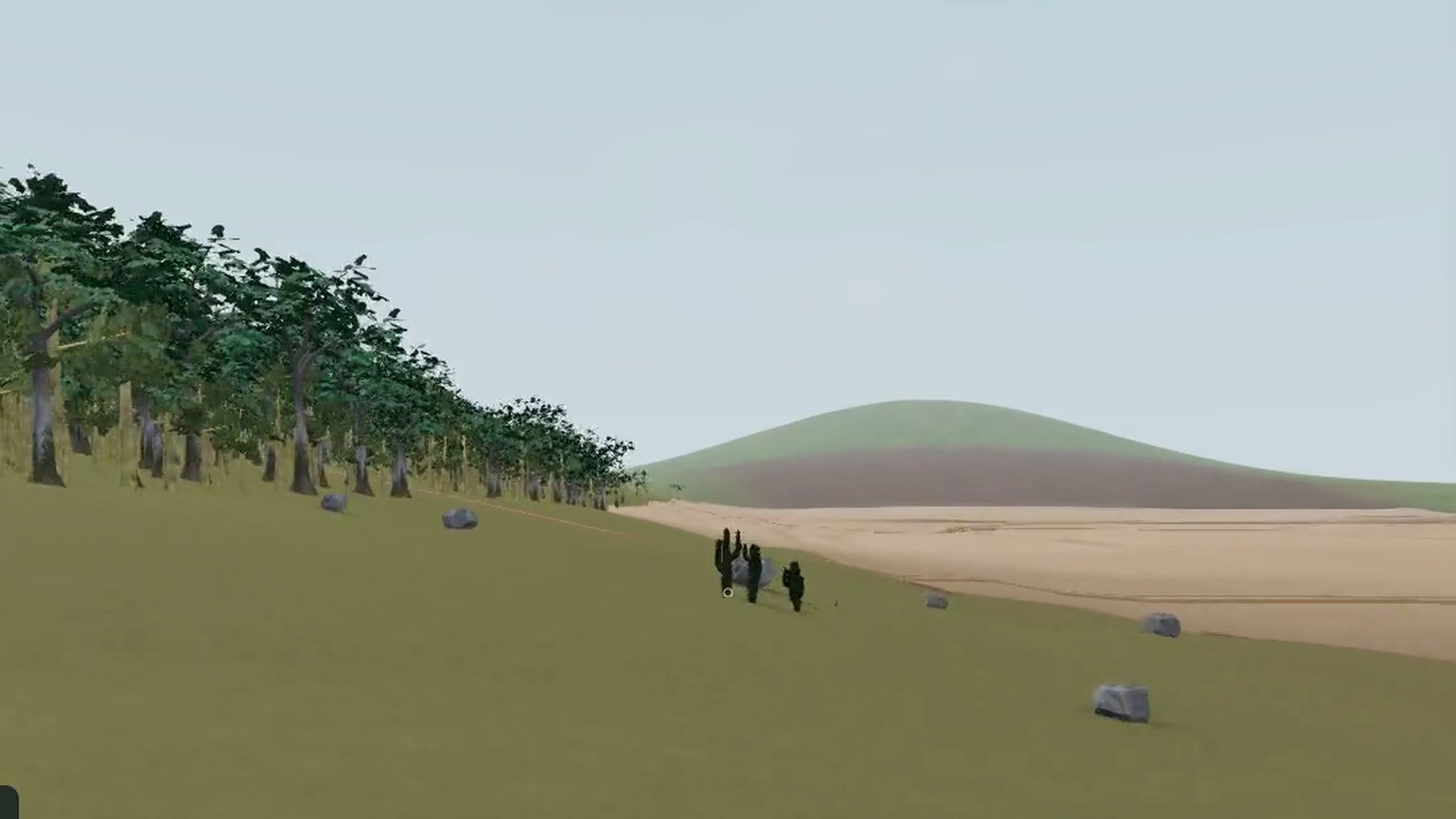}\\[-1mm]
{\small (d) Sheep interaction}&{\small (e) Farmland interaction}&{\small (f) Forest--desert transition}\\[1.5mm]
\includegraphics[width=.322\textwidth]{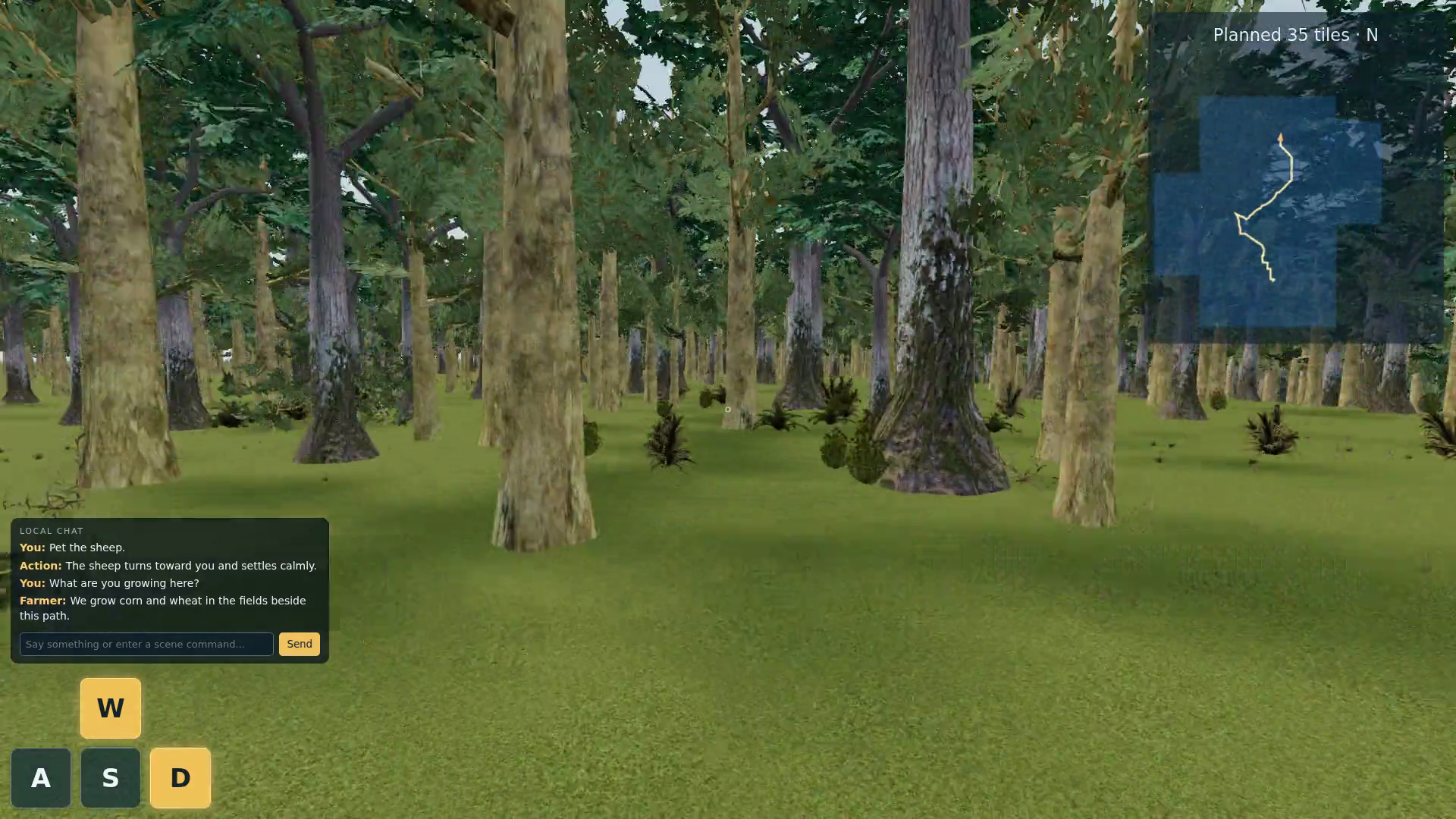}&
\includegraphics[width=.322\textwidth]{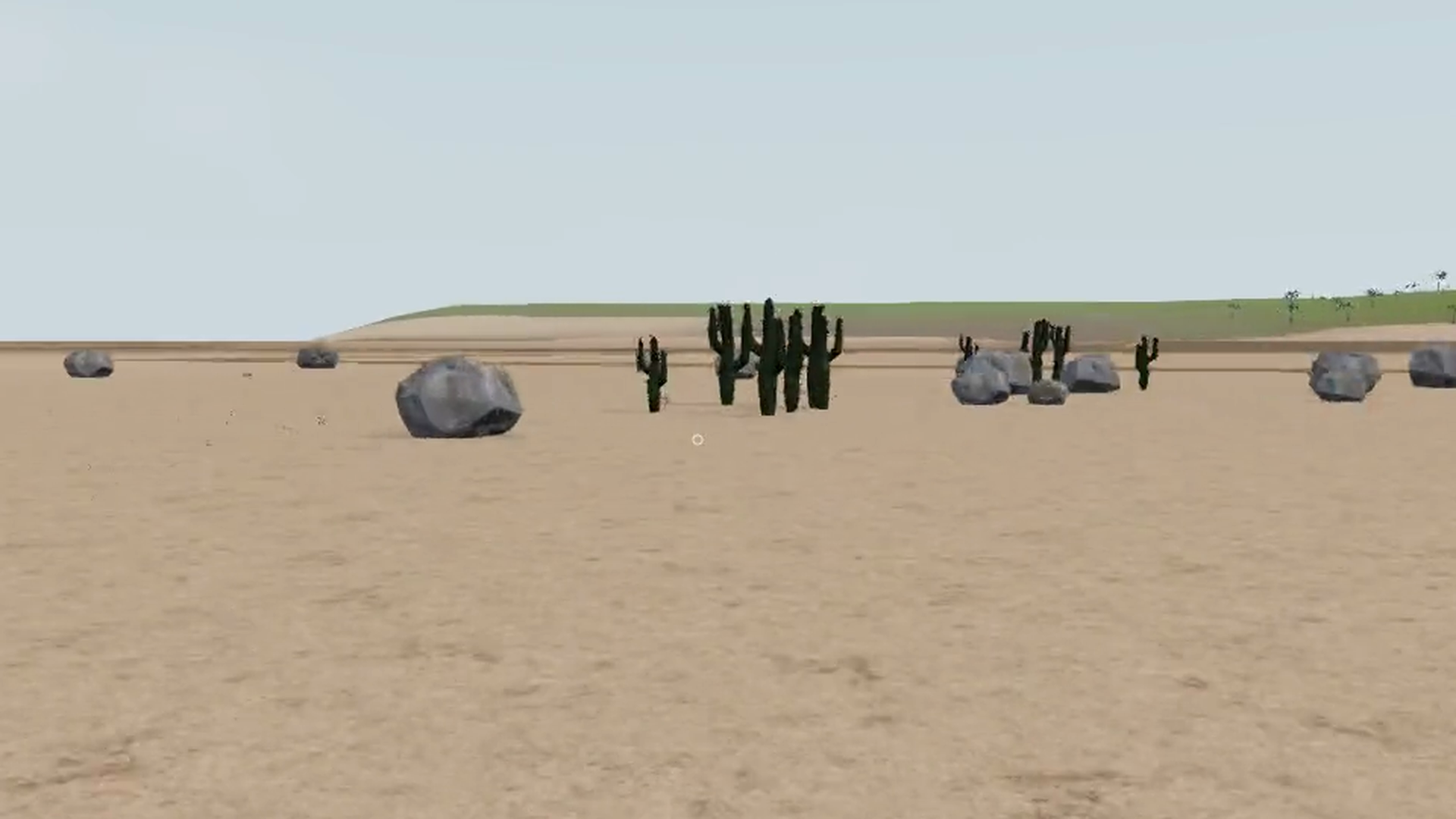}&
\includegraphics[width=.322\textwidth]{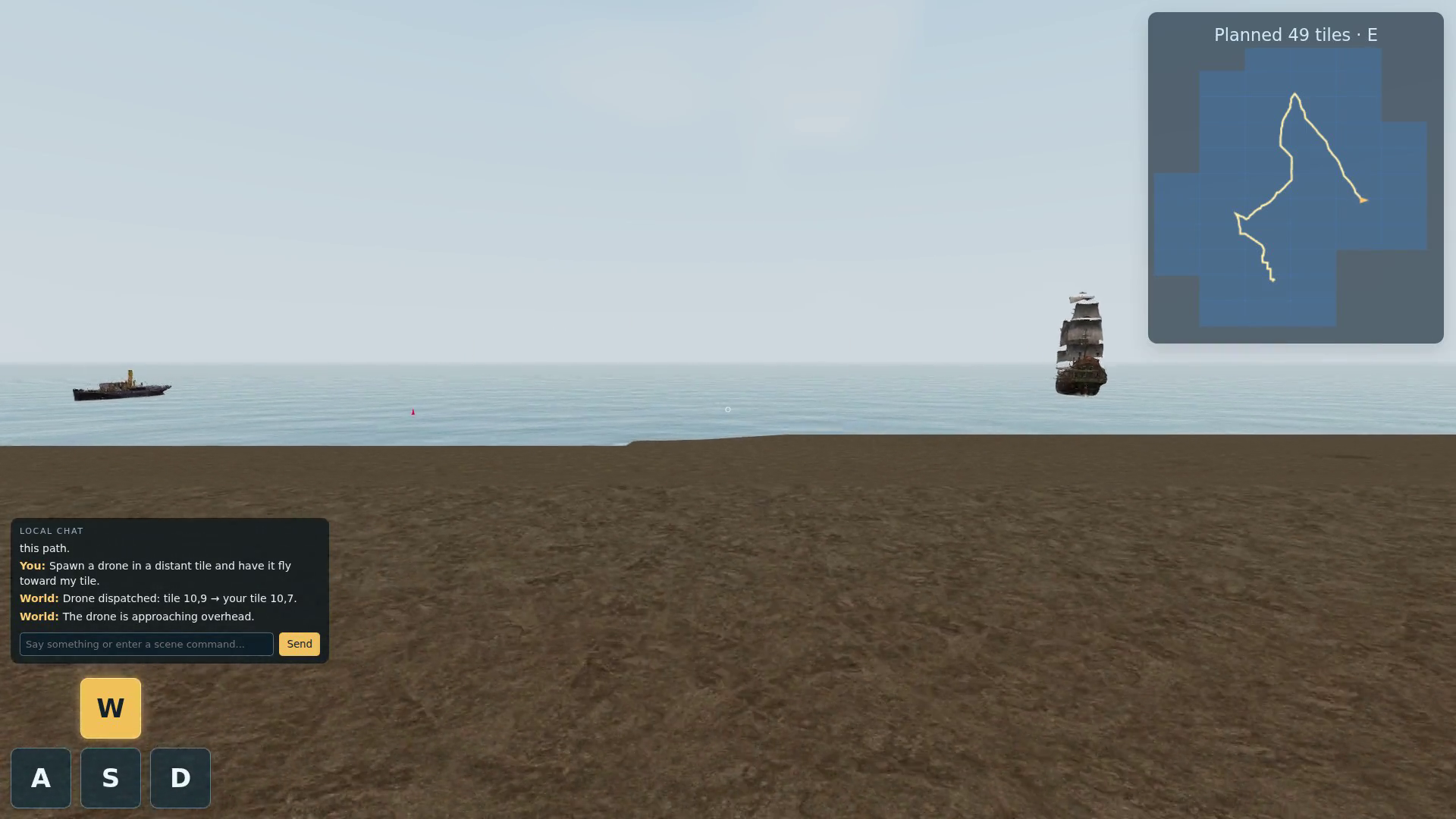}\\[-1mm]
{\small (g) Dense forest}&{\small (h) Desert}&{\small (i) Coast}
\end{tabular}
\caption{Sequential visual cases from one continuous WorldStream runtime tour recorded at 1920$\times$1080. The ordered samples show interaction and traversal through urban, rural, agricultural, forest, transition, desert, and coastal environments. Panels (a), (c), (e), (f), and (h) are crops of their source frames to make the recorded dialogue, rural dwelling, farmer and crops, forest--desert boundary, rocks, and cacti legible at three-column scale; the remaining panels show complete game-view frames.}
\label{fig:runtime-cases}
\end{figure}

Evaluation proceeds along four mechanisms in the runtime: hierarchical planning, movement-driven prefetch, grounded language interaction, and local execution. Figure~\ref{fig:runtime-cases} first shows representative outcomes from persisted runtime and interaction runs. Subsequent experiments isolate one mechanism at a time while holding its inputs and downstream implementation fixed.

\paragraph{Setup.} Rendering uses a Three.js-based WebGL runtime at $1920\times1080$. The executor runs three persistent compile worker processes, and demand is re-evaluated over a required $3\times3$ neighborhood. Reported times are wall-clock measurements unless stated otherwise. Three limits apply throughout. Planning ablations report one complete run per variant: they control the spatial input and the execution path, and do not estimate model-sampling variance. Prefetch replays measure backend tile readiness, with browser installation excluded. The interaction benchmark measures interpretation against fixed world facts, and does not isolate a complete edit--evict--revisit cycle.

\subsection{Planning Mechanisms}

\begin{figure}[!t]
\centering
\includegraphics[width=\textwidth]{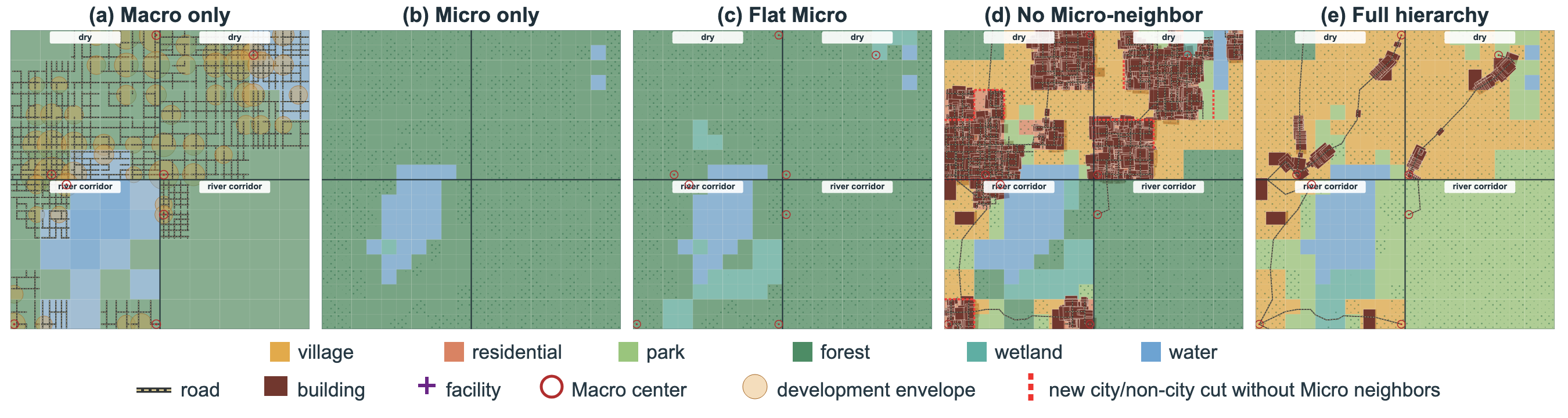}
\caption{Hierarchical planning ablation on the same four-Macro city-to-rural request. (a) Macro only provides settlement envelopes and a connected road skeleton but no local content. (b) Micro only uses accepted neighbors without a regional objective, retains natural land cover, and fails to form the requested settlement. (c) Flat Micro makes independent local decisions and lacks city-scale organization. (d) Without Micro-neighbor context, Macro centers and road ports remain, but tiles are repeatedly overbuilt and abrupt city/non-city cuts appear. (e) The Full hierarchy produces a connected, bounded urban-to-rural gradient.}
\label{fig:planner-ablation}
\vspace{.7em}
\captionof{table}{Compiled geometry for the locally realized variants in Figure~\ref{fig:planner-ablation}. Counts are total published instances. For variants with developed content, boundary jump is the mean absolute change in semantic development level across adjacent quadrant edges; lower is smoother. Macro only is omitted because it publishes no local content.}
\label{tab:planner-ablation}
\small
\renewcommand{\arraystretch}{1.08}
\setlength{\tabcolsep}{10pt}
\begin{tabular}{@{}lrrrr@{}}
\toprule
\textbf{Variant} & \textbf{Roads} & \textbf{Buildings} & \textbf{Facilities} & \textbf{Boundary jump}\\
\midrule
\textbf{Full hierarchy} & 47 & 250 & 13 & \metricgain{0.089}\\
\addlinespace[.12em]
\textbf{Micro only} & 0 & 0 & 0 & --\\
\textbf{Flat Micro} & 0 & 0 & 0 & --\\
\textbf{No Micro-neighbor} & 686 & 2,686 & 16 & 0.361\\
\bottomrule
\end{tabular}
\end{figure}

Planning is evaluated in two steps, beginning with the responsibilities of its spatial scales. All settings use the same city-to-rural request, geographic evidence, and local generators; only the information available to planning changes. Removing Macro context reveals the structure produced from local evidence alone, while removing accepted Micro neighbors reveals how regional intent behaves without local continuity. The resulting failures are complementary. Local planning without Macro context remains terrain-compatible but never forms the requested city. Macro guidance without neighbor context preserves centers and coarse roads, yet repeatedly overbuilds individual tiles and creates abrupt transitions. Figure~\ref{fig:planner-ablation} and Table~\ref{tab:planner-ablation} therefore expose a clear division of labor: Macro context determines where large-scale structure belongs, whereas Micro neighbors regulate how that structure continues across local boundaries.

With the spatial hierarchy established, the second planning experiment isolates the resolution step that makes its typed output executable. To remove model variation, 100 probability distributions produced by the decision model are retained and evaluated under two resolution rules. The independent baseline selects the highest-probability label for each quadrant separately. The joint solver uses the same probabilities but searches for an assignment that simultaneously satisfies terrain, water, biome, Macro-role, and adjacency constraints. Table~\ref{tab:typed-constraint-ablation} reports the fraction of assignments satisfying each constraint; overall success requires every hard constraint.

\begin{figure}[!t]
\centering
\captionof{table}{Overall and per-constraint success rates on shared model-produced probability distributions. Overall success requires all hard constraints; the final row reports multiplicative gain over independent selection.}
\label{tab:typed-constraint-ablation}
\small
\centering
\setlength{\tabcolsep}{2.8pt}
\begin{tabularx}{\textwidth}{@{}>{\raggedright\arraybackslash}p{2.35cm}*{7}{C}@{}}
\toprule
\textbf{Selection method} & \textbf{Overall} & \textbf{Biome} & \textbf{Terrain} & \textbf{Macro} & \textbf{Suburban} & \textbf{Landsc.} & \textbf{Water}\\
\midrule
\textbf{Independent} & 37\% & 55\% & 69\% & 74\% & 82\% & 84\% & 88\%\\
\textbf{Joint solver} & 100\% & 100\% & 100\% & 100\% & 100\% & 100\% & 100\%\\
\addlinespace[.15em]
\metricgain{Gain} & \metricgain{$\times$2.70} & \metricgain{$\times$1.82} & \metricgain{$\times$1.45} & \metricgain{$\times$1.35} & \metricgain{$\times$1.22} & \metricgain{$\times$1.19} & \metricgain{$\times$1.14}\\
\bottomrule
\end{tabularx}
\end{figure}

Independent selection leaves 63\% of plans invalid even though every field comes from the same model output. Joint resolution makes all 100 plans executable without changing those probabilities. This shows that structured output alone is insufficient: the fields must be resolved as a coupled spatial decision. Hierarchy and joint constraints thus address different failure modes---one organizes intent across space, while the other enforces consistency within each local plan.

\subsection{Streaming Readiness}

\begin{figure}[!t]
\centering
\includegraphics[width=\textwidth]{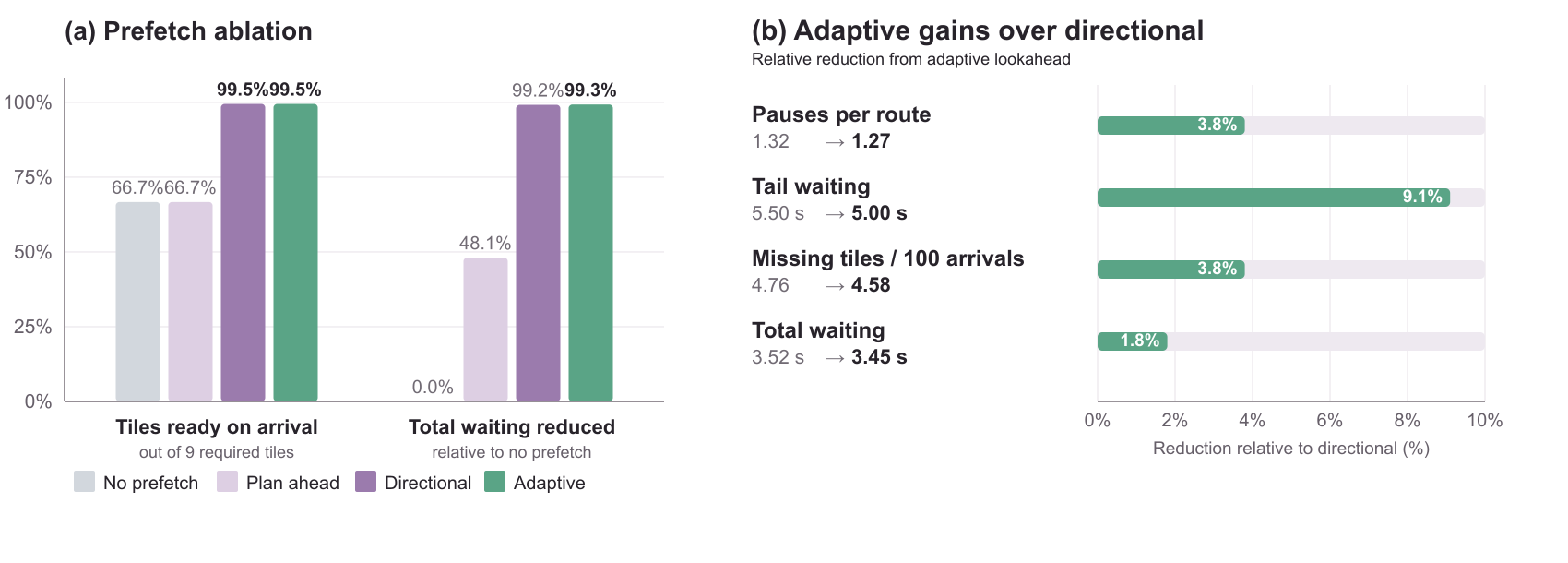}
\caption{Prefetch ablation over 100 route replays. (a) Four methods are compared on arrival coverage and 4~km waiting reduction; directional prefetch prepares almost the full $3\times3$ neighborhood before arrival. (b) Relative reductions obtained by adaptive lookahead over the directional policy on four lower-is-better metrics; raw directional and adaptive values are shown beside each metric.}
\label{fig:prefetch-ablation}
\end{figure}

Streaming readiness is evaluated by separating semantic lookahead from geometric readiness at tile arrival. The same 100 route traces are replayed with identical planning and construction times, worker limits, and a required $3\times3$ arrival neighborhood. The four controlled settings differ only in when their planning and construction stages become eligible. \emph{No prefetch} starts both stages only after a tile enters the current demand window. \emph{Plan ahead} requests semantic plans along the predicted route before arrival, but still defers construction until the tile enters the local construction window. \emph{Directional} also admits construction along the motion-predicted corridor, using fixed planning and construction horizons. \emph{Adaptive} retains the same directional policy but adjusts its lookahead from recent dispatch-to-finish times, expanding the active horizon when preparation becomes slower. We measure how many of the nine required tiles are ready at arrival and their cumulative waiting time on the complete 4~km benchmark route (Figure~\ref{fig:prefetch-ablation}).

Semantic readiness and visible readiness are not interchangeable. Planning ahead shortens recovery while leaving arrival coverage unchanged, as construction still begins too late. Directional prefetch is the mechanism that prevents most stalls: it brings almost the full neighborhood into the ready set before arrival and reduces total route waiting to 3.52~s. Adaptive lookahead then reduces the remaining tail. These mechanisms contribute differently: semantic lookahead reduces the cost of a miss, directional construction avoids the miss, and adaptation tunes the residual margin.

\subsection{Grounded Interaction}

Grounded interaction is measured by whether open-ended natural language selects the intended executable world operation. The same underlying world change can be phrased in many different ways, while scope, negation, compound instructions, and references to missing entities can make an otherwise plausible request unsafe or ambiguous. Each free-form request is therefore scored against either its intended world operation or safe non-execution when no valid operation exists. The benchmark uses 80 bilingual requests spanning actor and landmark interaction, current and future world editing, ambiguous or compound instructions, explicit no-action language, and unavailable targets or actions (Table~\ref{tab:interaction-summary}). World facts, retained state, and available runtime capabilities are held fixed across all requests, so measured success reflects language interpretation rather than variation in the underlying scene.

The intended operation is recovered in 76 of 80 requests (95.0\%). Performance is strongest for explicit actions and safety-sensitive cases: no-action requests and unavailable targets or capabilities are handled perfectly, without substituting a different entity or operation. The remaining errors concentrate on scope---especially distinguishing a future-world instruction from a current edit---rather than on recognizing the requested content itself. Temporal and spatial scope therefore need the same explicit representation as action and target.

\begin{table}[H]
\centering
\caption{Accuracy of mapping open-ended requests to the intended world operation or safe non-execution.}
\label{tab:interaction-summary}
\small
\renewcommand{\arraystretch}{1.08}
\setlength{\tabcolsep}{6pt}
\begin{tabularx}{\textwidth}{@{}>{\raggedright\arraybackslash}p{3.15cm}>{\centering\arraybackslash}p{1.45cm}Y@{}}
\toprule
\textbf{Request family} & \textbf{Correct} & \textbf{Tested distinction}\\
\midrule
\textbf{Actor / landmark} & \metricgain{19 / 20} & \tablekey{Action or direction} grounded to a \tablescope{known target}\\
\textbf{World editing} & \metricgain{18 / 20} & \tablescope{Present-region} \tablekey{edit} or \tablescope{future-region} \tablekey{intent}\\
\textbf{Clarification} & \metricgain{19 / 20} & \tablekey{Ambiguous scope} or multiple actions\\
\addlinespace[.2em]
\textbf{No action} & \metricgain{10 / 10} & \tablekey{Negation}, explanation, or quoted example\\
\textbf{Unavailable} & \metricgain{10 / 10} & \tablescope{Missing target} or \tablescope{unsupported operation}\\
\bottomrule
\end{tabularx}
\end{table}

\subsection{Local Execution Efficiency}

Local execution is evaluated by the cost of compiling an accepted plan into publishable tile records. The executor discards results whose source revision has become obsolete, while the optimization targets repeated work inside a single compilation. Terrain checks are shared across nearby placements instead of recomputed for every building or plant; reusable spatial indices accelerate collision and spacing tests; asset, ecology, and spatial-field data are cached or kept in compact form rather than repeatedly parsed and copied; and vegetation layers share deterministic cell-level sampling instead of initializing each candidate independently. Together, these changes reduce the cost of generating dense content while preserving every published object record exactly. Table~\ref{tab:compile-optimization} reports their combined effect across representative tile contents.

\begin{table}[htbp]
\centering
\caption{Wall-clock time for compiling accepted semantic plans into publishable tile records before and after the complete optimization. Both versions publish identical objects, asset identities, transforms, and instance counts. The final row reports compilation speedup.}
\label{tab:compile-optimization}
\small
\begin{tabularx}{\textwidth}{@{}l*{7}{C}@{}}
\toprule
 & \shortstack{\textbf{Dense}\\\textbf{forest}} & \shortstack{\textbf{Forest--}\\\textbf{wetland}} & \shortstack{\textbf{Light}\\\textbf{village}} & \textbf{Park} & \shortstack{\textbf{Light}\\\textbf{mountain}} & \textbf{Farmland} & \shortstack{\textbf{Coast--}\\\textbf{wetland}}\\
\midrule
\textbf{Before (s)} & 10.780 & 5.102 & 1.196 & 1.218 & 0.164 & 0.176 & 0.205\\
\textbf{After (s)} & 3.323 & 2.699 & 0.718 & 0.960 & 0.159 & 0.171 & 0.188\\
\addlinespace[.15em]
\metricgain{Speedup} & \metricgain{$\times$3.24} & \metricgain{$\times$1.89} & \metricgain{$\times$1.67} & \metricgain{$\times$1.27} & \metricgain{$\times$1.03} & \metricgain{$\times$1.03} & \metricgain{$\times$1.09}\\
\bottomrule
\end{tabularx}
\end{table}

Vegetation-heavy terrain shows the largest speedup: dense-forest compilation is $3.24\times$ faster and forest--wetland compilation is $1.89\times$ faster, while light-village and park tiles improve by $1.67\times$ and $1.27\times$, respectively. Across 12 newly published tiles in the complete streaming runtime, local compilation averages 2.134~s and reaches 3.318~s at maximum. Process-level concurrency contributes separately to runtime throughput: replacing two compile threads with two persistent worker processes raises warm throughput from 0.234 to 0.464 tiles/s, a measured 1.99$\times$ increase, because compiler state is then loaded once per process and independent tile compilations no longer contend for a single interpreter under the same workload and runtime configuration.

Taken together, the experiments trace one path from a semantic decision to a continuously inhabitable world. Regional planning establishes large-scale structure; directional scheduling converts that structure into geometry before the player arrives; retained state carries the accepted result into later interaction and reconstruction. The resulting world expands in explored area while its semantics, geometry, and accumulated user changes remain coupled over movement and time.

\section{Discussion and Conclusion}
\label{sec:discussion}

\subsection{Design Implications and Limitations}

At system level, persistent explicit state provides the interface among semantic planning, procedural construction, interaction, and rendering. Planning writes intent, generators realize that intent under spatial constraints, and the renderer derives a transient view. The retained record carries identity across all three stages. Related code-driven and programmable world models similarly separate executable state evolution from visual realization~\cite{chen2026codeworldmodel,huang2026programmable}; WorldStream extends this separation to movement-driven expansion through bounded residency, cross-tile commitments, revision-aware interaction, and deterministic revisit. This organization allows planners, generators, and renderers to evolve independently while preserving one world state.

The current prototype instantiates this organization with a bounded generation vocabulary, local controllers, and relational edit geometry. Curated geographic rules provide terrain and ecological plausibility, while reconstruction depends on retained records and compatible generator and asset versions. Expansion quality follows the available surface evidence, planning throughput, and readiness maintained ahead of movement. Long-running worlds also require storage policies for accumulated records and simulation policies for dynamic actors outside the active neighborhood. These boundaries define future questions while preserving planning, streaming, and persistence interfaces.

\subsection{Outlook and Conclusion}

Looking forward, the retained plan-and-record interface supports two complementary paths toward higher visual fidelity. A streaming visual decoder could consume low-poly realization, camera state, object identity, and edits, combining explicit control with long-horizon visual synthesis from WorldPlay and programmable or code-guided renderers~\cite{sun2025worldplay,huang2026programmable,li2026videococo}. Its evaluation would measure added latency and the consistency of edited places after revisit. Alternatively, richer explicit generators could replace the present assets and local realization procedures while preserving the same planner contracts. Agentic open-world construction, recursive scene programs, and unified urban context provide candidate mechanisms~\cite{guo2026worldclaw,li2026rcwm,huang2026holoworld}; prefetch and caching would prepare and reuse their detail within the existing streaming budgets.

\system{} formulates world generation as a continuous runtime. A typed decision model supplies connected regional and local semantics; deterministic local code realizes them under finite budgets; and retained tile records carry identity beyond resident geometry. Direction-aware scheduling prepares the world ahead of movement, while grounded language actions update the same records used by planning, construction, eviction, and revisit. Together these mechanisms connect model-based planning, explicit 3D generation, streaming exploration, interaction, and durable world identity within one extensible world-model loop.

\bibliographystyle{unsrtnat}
\bibliography{references}

\end{document}